\documentclass[10pt,twocolumn,letterpaper]{article}

\usepackage{cvpr}              % To produce the CAMERA-READY version
\usepackage{graphicx}
\usepackage{amsmath}
\usepackage{amssymb}
\usepackage{booktabs}

\usepackage{times}
\usepackage{epsfig}
\usepackage{graphicx}
\usepackage{amsmath}
\usepackage{amssymb}
\usepackage{subcaption}
\usepackage{stackrel}
\usepackage{wasysym}
\usepackage{dsfont}
\usepackage{enumitem}
\usepackage{bm}
\usepackage{booktabs}
\usepackage{multirow}
\setlist{nolistsep}
\usepackage[hang,flushmargin]{footmisc} 

\usepackage[dvipsnames]{xcolor}
\usepackage{tabu}
\usepackage{longtable}
\usepackage{color}
\usepackage{colortbl}
\usepackage{mathtools}
\usepackage{caption}%0321

\usepackage{lipsum}
\usepackage[misc]{ifsym}
 \usepackage{amsthm}
\theoremstyle{definition}

\usepackage{arydshln}

\usepackage{pifont}% http://ctan.org/pkg/pifont
\newcommand{\cmark}{\ding{51}}%
\newcommand{\xmark}{\ding{55}}%

\usepackage[dvipsnames]{xcolor}

\definecolor{mypink1}{rgb}{0.858, 0.188, 0.478}
\definecolor{mypink2}{RGB}{219, 48, 122}
\definecolor{mypink3}{cmyk}{0, 0.7808, 0.4429, 0.1412}
\definecolor{mygray}{gray}{0.6}

\newtheorem{corollary}{Corollary}
\newtheorem{proposition}{Proposition}

\usepackage{titletoc}
\usepackage{float}

\usepackage[resetlabels]{multibib}
\newcommand{\bibnamec}{References}
\newcites{supp}{\bibnamec}
\usepackage[title]{appendix}

\newcommand{\tabincell}[2]{\begin{tabular}{@{}#1@{}}#2\end{tabular}}

\usepackage[pagebackref,breaklinks,colorlinks]{hyperref}

\usepackage[capitalize]{cleveref}
\crefname{section}{Sec.}{Secs.}
\Crefname{section}{Section}{Sections}
\Crefname{table}{Table}{Tables}
\crefname{table}{Tab.}{Tabs.}

\def\cvprPaperID{1273} % *** Enter the CVPR Paper ID here
\def\confName{CVPR}
\def\confYear{2022}

\begin{document}

%%%%%%%%% TITLE
\title{SoT: Delving Deeper into Classification Head for Transformer}

% \author{Jiangtao Xie, Ruiren Zeng\corref{cor1}}
% \cortext[cor1]{hhhh}
% \author{Ruiren Zeng}
\author{Jiangtao Xie$^{1}$\footnotemark[3], Ruiren Zeng$^{1}$\footnotemark[3], Qilong Wang$^{2}$, Ziqi Zhou$^{3}$, Peihua Li$^{1}$\footnotemark[4]\\
$^{1}$ Dalian University of Technology, $^{2}$ Tianjin University, $^{3}$ MEGVII Technology
%{\tt\small peihuali@dlut.edu.cn}
}
% \author{Jiangtao Xie$^{1}$, Ruiren Zeng$^{1}$,}
% \author{Qilong Wang$^{2}$,}
% \author{Ziqi Zhou$^{3}$,}
% \author{Peihua Li$^{1}$\\
% $^{1}$ Dalian University of Technology, $^{2}$ Tianjin University, $^{3}$ MEGVII Technology\\
% {\tt\small peihuali@dlut.edu.cn}
% }
%\author{Peihua Li$^{\dagger}$\\
%\and

\maketitle
\thispagestyle{empty}

\renewcommand{\thefootnote}{\fnsymbol{footnote}} %将脚注符号设置为fnsymbol类型，即特殊符号表示
\footnotetext[3]{These authors contributed equally to this work.} %对应脚注[1]
\footnotetext[4]{Corresponding author. Email: peihuali@dlut.edu.cn} %对应脚注[1]

%%%%%%%%% ABSTRACT
\begin{abstract}
	
Transformer models are not only successful in natural language processing (NLP)  but also demonstrate high potential in computer vision (CV).  Despite great advance, most of works only focus on improvement of  architectures but pay little attention  to the classification head. For years  transformer models base exclusively on classification token to construct the final classifier, without explicitly  harnessing high-level word tokens.  In this paper, we propose a novel  transformer model called second-order transformer (SoT),  exploiting simultaneously the classification token and word tokens for the classifier.  Specifically, we empirically disclose that high-level word tokens contain rich information, which per se are very competent with the classifier and moreover, are  complementary to the classification token. To effectively harness such rich information, we propose  multi-headed global cross-covariance pooling with singular value power normalization, which shares similar philosophy and thus is compatible with the transformer block, better than commonly used pooling methods. Then, we study comprehensively how to explicitly combine word tokens with classification token for building the final classification head. For CV tasks, our SoT significantly improves state-of-the-art vision  transformers on  challenging benchmarks including  ImageNet and ImageNet-A. For NLP tasks, through fine-tuning based on pretrained language transformers including GPT and BERT, our SoT greatly boosts the performance on widely used tasks such as CoLA and RTE.  Code will be available at \href{https://peihuali.org/SoT}{https://peihuali.org/SoT}.  

\end{abstract}

%%%%%%%%% BODY TEXT
\section*{1.~Introduction}

\begin{table}[t]
	\centering
	\footnotesize
	\setlength{\tabcolsep}{1.0pt}
	\renewcommand\arraystretch{1.2}
	
	\begin{minipage}{1\linewidth}
		\begin{subtable}{1\linewidth}
			\centering
			\begin{tabular}{l|c|c|c|c|c|c}
				\hline
				\multirow{2}{*}{\tabincell{c}{Vision  \\Transformer}}&\multicolumn{2}{c|}{DeiT-T~\cite{DeiT_ICML}}& \multicolumn{2}{c|}{T2T-7~\cite{T2T_ICCV21}}& \multicolumn{2}{c}{T2T-14 ~\cite{T2T_ICCV21}}\\
				\cline{2-7}
				&	IN   &  IN-A & IN & IN-A & IN & IN-A \\
				\hline
				ClassT& 72.2\;\;\;\;\;\;	& 7.3\;\;\;\;\;\;\;\;\; &  71.7\;\;\;\;\;\;  &  6.1\;\;\;\;\;\;	& 81.5\;\;\;\;\;\;	&	23.9\;\;\;\;\;\;\;\\
				
				WordT&77.9$_{5.7\uparrow}$ & 15.5$_{8.2\uparrow}$\;& 73.7$_{2.0\uparrow}$	& 7.8$_{1.7\uparrow}$	&82.1$_{0.6\uparrow}$& 26.6	$_{2.7\uparrow}$\\
				
				ClassT+WordT& 78.6$_{6.4\uparrow}$ & 17.5$_{10.2\uparrow}$& 74.5$_{2.8\uparrow}$	& 8.1$_{2.0\uparrow}$	&  82.6$_{1.1\uparrow}$ & 27.1$_{3.2\uparrow}$ \\
				\hline
				\hline
				\multirow{2}{*}{\tabincell{c}{Language \\Transformer}}&\multicolumn{2}{c|}{GPT~\cite{GPT-1}}& \multicolumn{2}{c|}{BERT-base~\cite{DBLP:conf/naacl/DevlinCLT19}}& \multicolumn{2}{c}{BERT-large~\cite{DBLP:conf/naacl/DevlinCLT19}}\\
				\cline{2-7}
				&	    CoLA & \tabincell{l}{\centering RTE}  & CoLA  & RTE   & CoLA & RTE \\
				\hline
				ClassT& 54.3\;\;\;\;\;\;& 63.2\;\;\;\;\;\;& 54.8\;\;\;\;\;\;&67.2\;\;\;\;\;\;&60.6\;\;\;\;\;\;&73.7\;\;\;\;\;\;\\
				
				WordT&	56.1$_{1.8\uparrow}$&65.0$_{1.8\uparrow}$   & 56.4$_{1.6\uparrow}$	&69.0$_{1.8\uparrow}$    &61.4$_{0.8\uparrow}$	&   74.4$_{0.7\uparrow}$  \\
				
				ClassT+WordT&	57.3$_{3.0\uparrow}$& 65.4$_{2.2\uparrow}$& 58.0$_{3.2\uparrow}$	&69.3$_{2.1\uparrow}$  &61.8$_{1.2\uparrow}$  &75.1$_{1.4\uparrow}$ \\
				\hline
				
			\end{tabular}%
		\end{subtable}
	\end{minipage}
	\caption{Accuracies  (\%) of  transformer models which  use single classification token (ClassT), single word tokens (WordT) and their combination (ClassT+WordT).  We showcase performance of vision  transformers (i.e., DeiT and T2T) on ImageNet (IN)~\cite{imagenet_cvpr09} and ImageNet-A (IN-A)~\cite{ImageNet-A},  and that of language transformers (i.e., GPT and BERT) on CoLA~\cite{warstadt2018neural} and RTE~\cite{bentivogli2009fifth}. The results  indicate  word tokens per se are very competent with classifier and moreover, are complementary to the classification token. See Sec.~\hyperref[section:experiments]{\textcolor{red}{4}} for details and  for diverse results. }
	\label{table: ClassT or WordT and fusion}	
\vspace{-6pt}	
\end{table}

In the past years  transformer models have been very successful in the field of natural language processing (NLP)~\cite{NIPS2017_3f5ee243,DBLP:conf/naacl/DevlinCLT19,GPT-1}. The transformer architecture, solely based  on attention mechanisms, can naturally model long-range dependency of tokens and learn contextual knowledge. In  contrast, the convolutional neural networks (CNNs)~\cite{LeNet1989,nips2012cnn} lack such capability as the  convolutions are fundamentally local operations.  The success of transformer models has attracted great interests of computer vision (CV) researchers~\cite{wang2018non,LambdaNetworks-ICLR2020,DBLP:journals/corr/abs-2012-12556,DBLP:journals/corr/abs-2101-01169}. Recently, a vision  transformer model called ViT~\cite{ViT} is proposed entirely based on a stack of transformer blocks, which has matched or outperformed state-of-the-art CNNs when pre-trained on ultra large-scale datasets of ImageNet-21K~\cite{imagenet_cvpr09} or JFT-300M~\cite{JFT-300M}. Thereafter, a number of pure transformer models, e.g., ~\cite{T2T_ICCV21,PSViT_ICCV21,PVT_ICCV21},  have been proposed to improve the architecture of transformers,  reporting impressive  performance gains when trained from scratch on ImageNet with 1K classes~\cite{imagenet_cvpr09}.

Despite great advance, for years  pure transformer models invariably build  final classifier  exclusively based on classification token, without explicitly harnessing high-level word tokens.  Although the classification token interacts with all word tokens$\,$\footnote{For brevity, we   use the notation of  ``word'' token for both NLP  and CV tasks, which, for the latter case, indicates image patch.} through the attention mechanisms across the network backbone, we conjecture the high-level word tokens by themselves contain rich information that the classification token fails to accommodate. Therefore,  exploiting only the classification token but excluding the word tokens from the classifier limits transformer models. Actually, we empirically find that rich information inherent in word tokens per se is very competent with classifier and moreover,  is complementary to the classification token. 

As show in Tab.~\ref{table: ClassT or WordT and fusion}, the experimental results on both CV and NLP tasks show that classification head solely based on word tokens (i.e., WordT) is often better than that based on single classification token (i.e., ClassT), and combination of WordT and ClassT further  boosts  the classification accuracy. 
Based on the empirical observations above, we rethink classification head for transformer, and propose a novel  transformer model, namely second-order transformer (SoT), to exploit simultaneously classification token and word tokens for the final classifier.  To this end,  there exist two key issues to be tackled: (1) how to effectively  aggregate the word tokens to fully explore their rich information; (2) how to explicitly combine word tokens with classification token for building the final classification head.

For effective token aggregation, we propose a  multi-headed global cross-covariance pooling (MGCrP) method, which  learns  a group of second-order, cross-covariance representations. Previous works~\cite{LiXWZ17,PAMI_2020_MPN-COV} have shown that structured normalization plays an important role for the second-order representations. Unfortunately,  existing structured normalization methods are not applicable to our MGCrP as it produces  asymmetric matrices. Therefore, we present a singular value power normalization (svPN) method in light of the  overall statistical analysis of data; moreover, an approximate, fast svPN algorithm is developed for guaranteeing  efficiency.   MGCrP shares similar philosophy and so is more consistent with the transformer block, clearly better than  global average pooling (GAP)~\cite{iclr2014_NIN,Szegedy_2015_CVPR,He_2016_CVPR} and global covariance pooling (GCP)~\cite{pami/LinRM18,PAMI_2020_MPN-COV} widely used in CNNs.

To build the classification head by combining word tokens with classification token, we propose three early fusion schemes, which combine the two kinds of token representations   through operations such as concatenation and sum,  along with one late fusion scheme  which integrates individual classification scores. These schemes are illustrated in Fig.~\ref{fig:fusion schemes} and their comparisons are given in Tab.~\ref{tab:comparison of fusion method}; among them, the sum scheme performs best.  Note that the proposed classification head is architecture-agnostic, which can be seamlessly integrated into a wide variety of vision  transformers and language transformers.  To verify the effectiveness of our SoT, we conduct experiments on both CV and NLP tasks.

Our contributions are summarized as follows.
\begin{itemize}
	\item  We dig into the transformer's classification head, finding that word tokens per se are very competent with classifier and moreover, is complementary to classification token. Based on this,  we propose a second-order transformer (SoT) model. As far as we know, our SoT makes the first attempt to exploit simultaneously classification token and word tokens for classification, which can span and benefit both CV and NLP tasks.
	\item  We propose multi-headed global  cross-covariance pooling  with structured normalization for mining word tokens, while  systemically studying several schemes to combine word tokens with classification token. As a result, we achieve a novel classification head, which is very effective and suitable for a variety of transformer architectures.
	\item  We perform thorough ablation study to validate our SoT. Extensive experiments show our SoT can significantly improve vision  transformers on challenging ImageNet and ImageNet-A. Meanwhile, our SoT is  very helpful to language transformers such as BERT and GPT, performing  better than its conventional counterpart on General Language Understating tasks.
\end{itemize}

\section*{2.~Related Works}\label{section-related work}

\vspace{4pt}\noindent \textbf{Transformer architectures} Solely based  on attention mechanisms without any recurrence and convolutions, the transformer models~\cite{NIPS2017_3f5ee243} can naturally model long-range dependencies and global context, outperforming LSTM~\cite{hochreiter1997long} and CNN~\cite{10.5555/3305381.3305510}  on NLP tasks. The  transformer models  pre-trained  on large-scale unlabeled corpus (e.g., GPT~\cite{GPT-1,GPT-2} and BERT~\cite{DBLP:conf/naacl/DevlinCLT19}) are able to learn powerful  language representations, and, after fine-tuning, achieve state-of-the-art results for a wide range of downstream language tasks~\cite{GPT-1,GPT-2,DBLP:conf/naacl/DevlinCLT19,roberta}.  For CV tasks,   Dosovitskiy  et al.~\cite{ViT} present a pure transformer architecture (i.e., ViT)  which  reports very promising performance when pre-trained on ultra large-scale datasets.  The works that follow greatly improve over ViT.  DeiT~\cite{DeiT_ICML} proposes an extra distillation token to transfer  knowledge from teacher models.  T2T-ViT~\cite{T2T_ICCV21} and PS-ViT~\cite{PSViT_ICCV21} focus on better  tokenization of vision  patches.   Swin Transformer~\cite{Swin_ICCV21} and  PVT-T~\cite{PVT_ICCV21} introduces hierarchical structure into the transformer. Conformer~\cite{Conformer_ICCV21} is a dual architecture which combines CNN with transformer. 
The pure transformer architectures, either in NLP or CV field, only use the classification token for the classifier, limiting the performance of models. As such, we propose a novel classification head, which exploits simultaneously the classification token and word tokens. 

\vspace{4pt}\noindent \textbf{Second-order pooling} GCP, also known as bilinear pooling, often produces  symmetric positive definite (SPD) matrices as global representations~\cite{Ionescu_2015_ICCV,lin2015bilinear}. It can capture second-order relations, performing much better than  GAP which only estimates first-order statistics~\cite{pami/LinRM18,PAMI_2020_MPN-COV}. Research has shown that normalization  techniques are very helpful to the second-order representations~\cite{lin2015bilinear,RANK-1-2020,LiXWZ17,Ionescu_2015_ICCV}.  Element-wise normalization improves GCP but fails to consider holistic structure of data~\cite{lin2015bilinear}.  The structured normalization such as matrix power normalization (MPN)~\cite{LiXWZ17,PAMI_2020_MPN-COV,lin2017improved} exploits the overall statistical structure and  geometric structure of covariance matrices,  and has greatly benefited  GCP.   As MPN depends on GPU-unfriendly  eigen-decomposition, iSQRT~\cite{LiXWG18} proposes a fast  method for computing matrix square root,  suitable for parallel implementation on GPU.  A recent study~\cite{WQL_What_CVPR20} has shown that  GCP with MPN improves Lipschitzness of the loss function, leading to fast convergence and robustness to distorted images. Different from  GCP, we propose multi-headed global cross-covariance pooling, which shares similar philosophy and is  consistent with transformers; furthermore, we propose a new structured method to normalize cross-covariance matrices.

%------------------------------------------------------------------------
\section*{3.~Second-order  Transformer  (SoT)}\label{section:So-ViT model}

In this section, we first describe  the SoT network, followed by the proposed multi-headed global cross-covariance pooling method.  Finally,  we introduce the normalization method for cross-covariance matrices.

\subsection*{3.1.~SoT Network}\label{subsection:SoT network}

\begin{figure}[t]
  	\begin{center}
  		\includegraphics[width=0.8\linewidth]{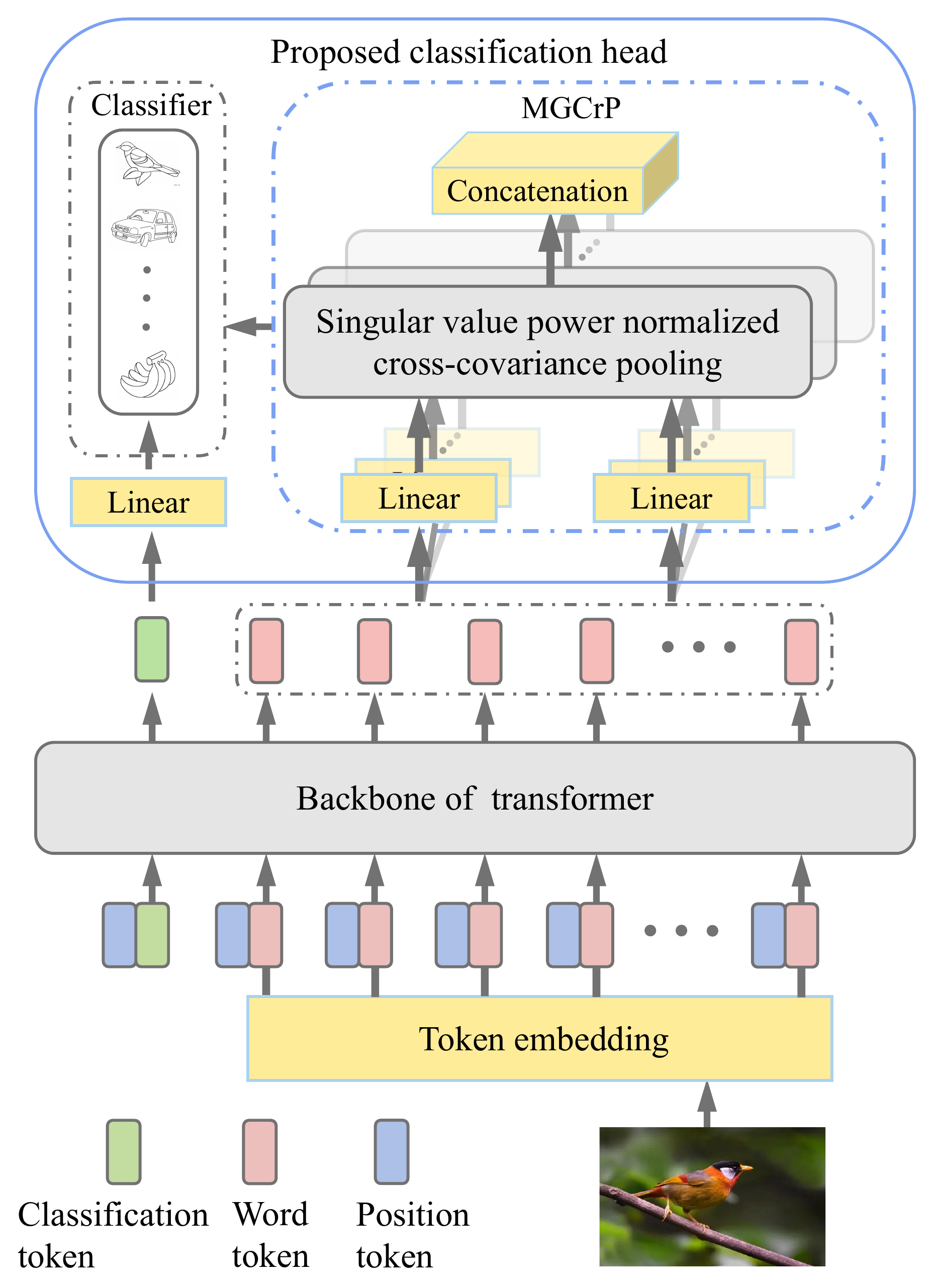}
  	\end{center}
  	\caption{Diagram of our SoT network. Given an input image, the token embedding module produces a sequence of word tokens, which is then prepended by a classification token. After added   by position tokens,  the sequence of tokens is subject to the backbone  consisting of a stack of standard transformer blocks. Finally, the classification token and the word tokens output from the backbone are fed to the proposed classification head.}
  	\label{fig:overview}
  	\vspace{-6pt}
  \end{figure}

    We illustrate  our SoT network in Fig.~\ref{fig:overview}.  Similar to~\cite{PSViT_ICCV21}, we develop a small, hierarchical module for  token embedding, which, based on off-the-shelf convolutions, gradually reduces spatial size of feature maps.  In designing the token embedding module, we evaluate varying type of  convolution blocks including ResNet block~\cite{He_2016_CVPR}, Inception block~\cite{Szegedy_2015_CVPR}, DenseNet block~\cite{Huang_2017_CVPR} and Non-local block~\cite{wang2018non}. This token embedding method  has capability to model  local image properties, which ViT~\cite{ViT} lacks  due to its naive embedding method. 
    
    The  features produced by the token embedding module are reshaped to a sequence of vectors as word tokens. 
    As in~\cite{ViT}, we prepend a learnable classification token  to the word token sequence,  and then  position embeddings are added. The modified token  sequence  is then fed to the backbone network, which consists of a stack of  standard transformer blocks, each containing a  multi-head self-attention   and a multi-layer perception.
    
    For  classification head, unlike the conventional transformers,  we explicitly combine the word tokens with the classification token. We investigate different  fusion methods to combine  two kinds of tokens, finding the sum scheme performs best which, therefore, is adopted in our SoT.  We design a family of transformers  of varying depths,  including a 12-layer SoT-Tiny, a 14-layer SoT-Small and a 24-layer SoT-Base; in addition, we design a 7-layer SoT for the sake of ablation study.  The details on these models  as well as on the token embedding module are provided  in the supplement~\ref{suppsection:SoT network}. 
    
\subsection*{3.2.~Proposed Classification Head}\label{subsection:cross-cov our paradgim}

The conventional pure transformer models only use classification token for the final classifier. As high-level word tokens   contain rich   information, neglect of them leads to information loss.  So we propose to combine the word tokens with the classification token for the final  classifier.

\begin{figure}[t!]
	%\centering
	\begin{subfigure}[b]{0.22\textwidth}
		\centering
		\includegraphics[height=1.4in]{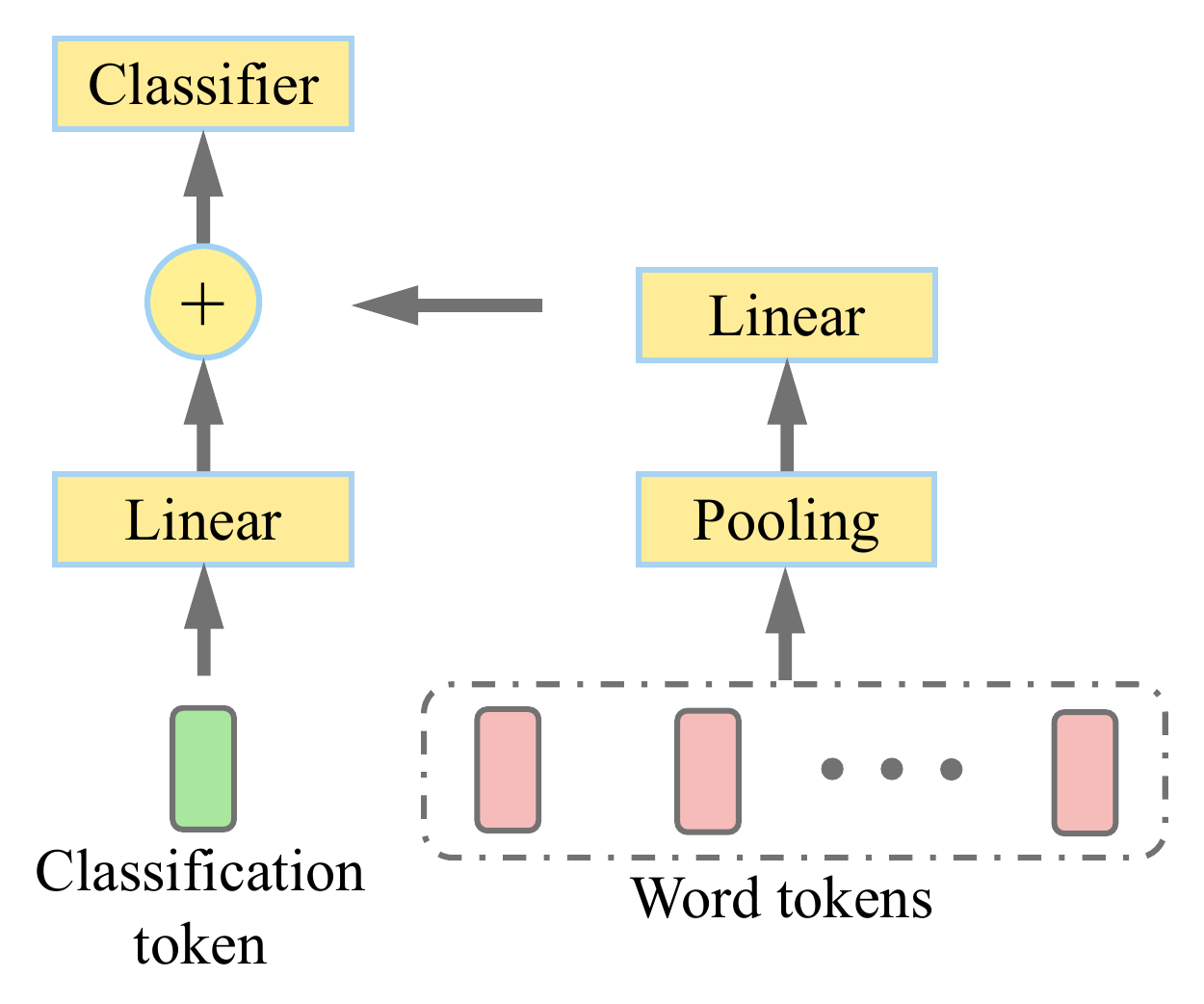}
		\caption{Sum }
		\label{subfigure:fusion_sum}
	\end{subfigure}%
	~\hspace{3pt}
	\begin{subfigure}[b]{0.22\textwidth}
		\centering
		\includegraphics[height=1.4in]{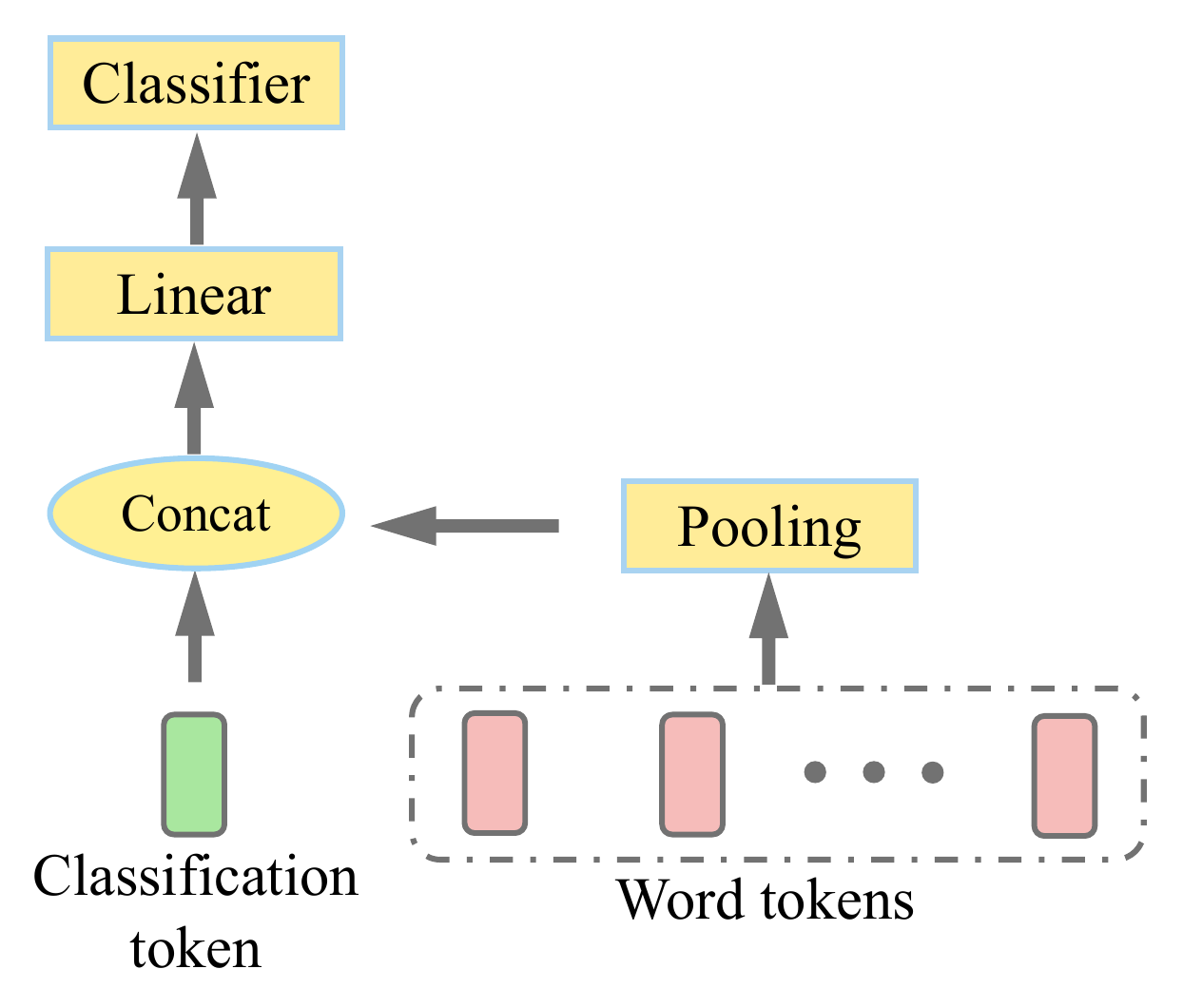}
		\caption{Concat}
		\label{subfigure:fusion_concat}
	\end{subfigure}
	
	\begin{subfigure}[b]{0.22\textwidth}
		\centering
		\includegraphics[height=1.4in]{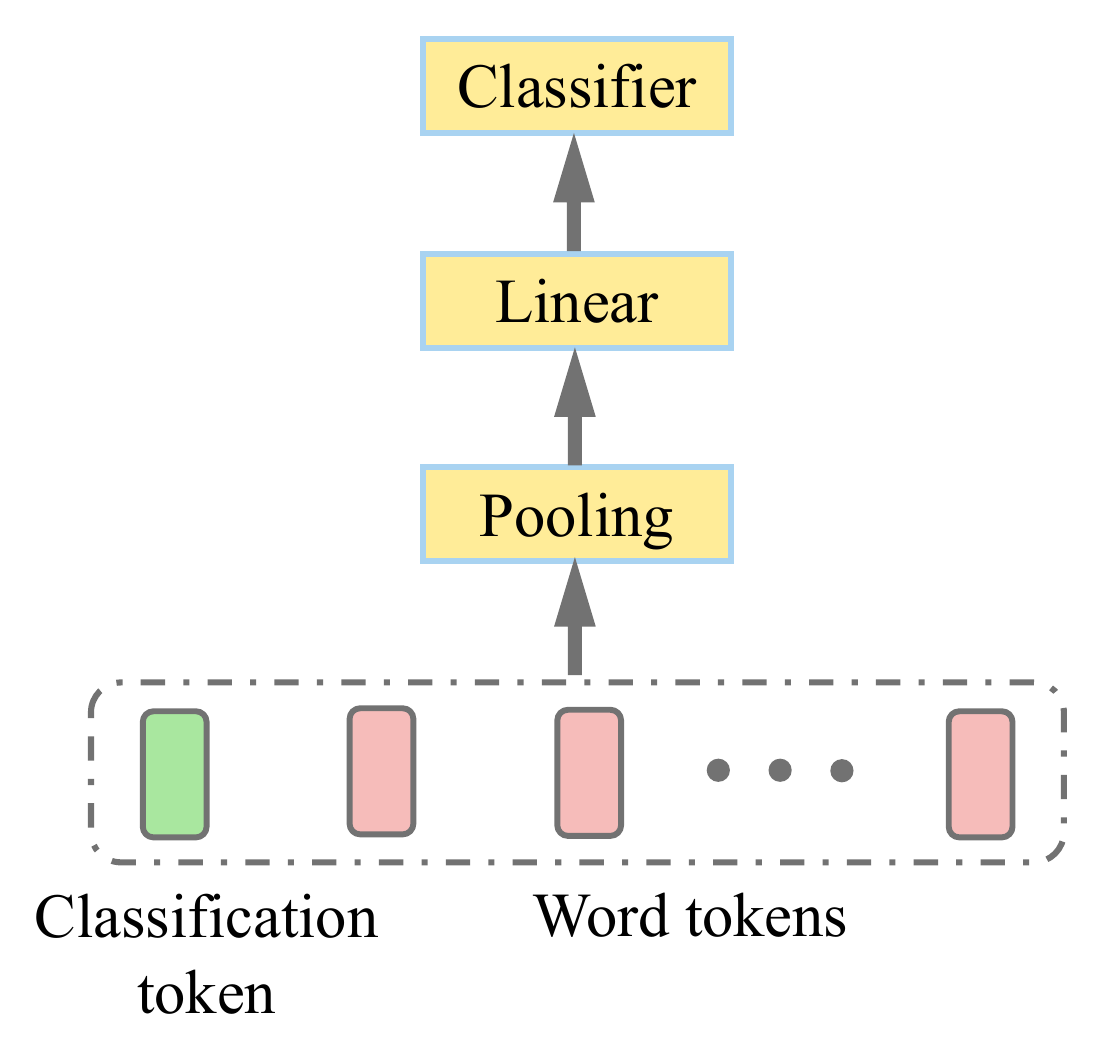}
		\caption{Aggr\_all}
		\label{subfigure:fusion_aggr_all}
	\end{subfigure}%
	~\hspace{3pt}
	\begin{subfigure}[b]{0.22\textwidth}
		\centering
		\includegraphics[height=1.4in]{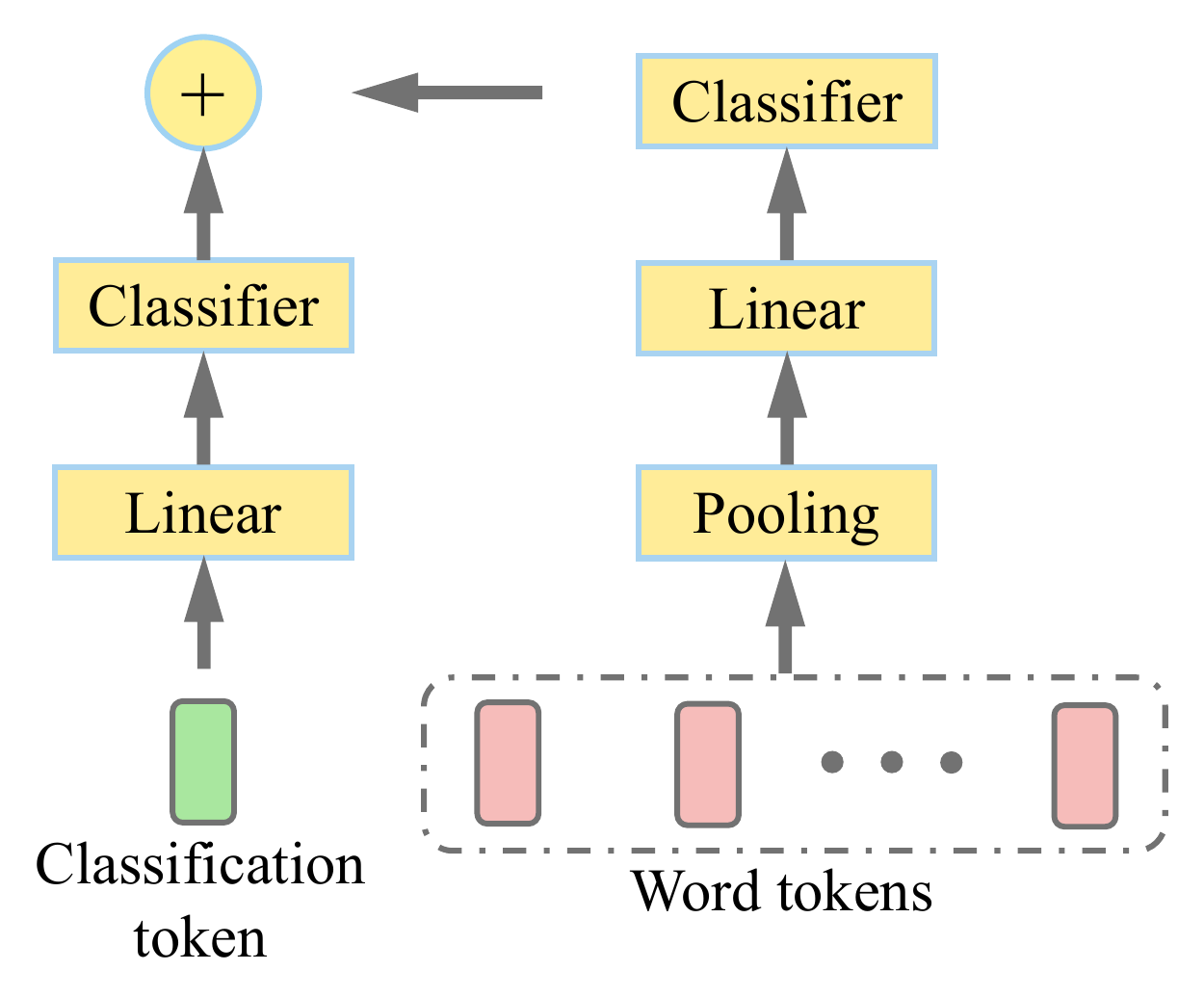}
		\caption{Late}
		\label{subfigure:fusion_late}
	\end{subfigure}	
	\caption{Fusion schemes designed for our classification head.}
	\label{fig:fusion schemes}
\end{figure}

We introduce three early  fusion schemes (i.e., $\mathrm{sum}$, $\mathrm{concat}$ and $\mathrm{aggr\_all}$) and a late fusion scheme (i.e., $\mathrm{late}$). Fig.~\ref{fig:fusion schemes} illustrates these schemes, where  a $\mathrm{Linear}$ transformation is equivalent to (and thus denoted by) a fully-connected (FC) layer.  For the $\mathrm{sum}$ scheme,  the classification token and aggregated word tokens are separately connected to  a FC layer and  are then summed,  before fed to the softmax classifier.  In the  $\mathrm{concat}$ scheme, the representations of classification token and the aggregated word tokens are  concatenated, followed by a FC layer and then a softmax classifier.  For the $\mathrm{aggr\_all}$ scheme, we directly aggregate all tokens including both the classification token and the word  tokens, and then connect the resulting representation to a  FC layer succeeded by  a softmax classifier.    In the late fusion scheme, the classification token and the word tokens are independently attached to a FC layer and a softmax classifier, and finally the two classification scores are added.

Let $\{\mathbf{z}_{i}\in \mathbb{R}^{p}, i=0,1,\ldots,q\}$ denote the features  of all tokens, where $\mathbf{z}_{0}$ indicates the classification token while the remaining ones indicate word tokens.  We concatenate the features of  word tokens to form a matrix  $\mathbf{Z}=[\mathbf{z}_{1},\mathbf{z}_{2},\cdots,\mathbf{z}_{q}] \in  \mathbb{R}^{p\times q}$.  The four fusion schemes can be formulated as
\begin{align}
\mathrm{sum}:\;\;&\mathrm{softmax}\big(\mathrm{FC}(\mathbf{z}_{0})+\mathrm{FC}(\mathrm{pool}(\mathbf{Z})\big)\\
\mathrm{concat}:\;\;&\mathrm{softmax}\left(\mathrm{FC}\left(\big[\mathbf{z}_{0},\mathrm{pool}(\mathbf{Z})\big]\right)\right)\nonumber\\
\mathrm{aggr\_all}:\;\;&\mathrm{softmax}\big(\mathrm{FC}(\mathrm{pool}([\mathbf{z}_{0},\mathbf{Z}])\big)\nonumber\\
\mathrm{late}:\;\;&\mathrm{softmax}\big(\mathrm{FC}(\mathbf{z}_{0})\big)+\mathrm{softmax}\big(\mathrm{FC}(\mathrm{pool}(\mathbf{Z}))\big)\nonumber
\end{align}
Here $\mathrm{pool}(\cdot)$ denotes a pooling function. The commonly used pooling functions are  global average pooling (GAP) and Global covariance pooling (GCP), which, however, are developed for  CNN architecture and may be sub-optimal  for the transformer. As described in~\cite[Sec. 3.2.2]{NIPS2017_3f5ee243}, the multi-head structure in the transformer block facilitates modeling information from different representation space. Inspired by this, we propose multi-headed global cross-covariance pooling (MGCrP)  with structured normalization, as illustrated in Fig.~\ref{fig:overview} (top-right).

\begin{table}[t]
	\centering
	\footnotesize
	\setlength{\tabcolsep}{2pt}
	\renewcommand\arraystretch{1.2}
	\begin{minipage}{1\linewidth}
		\begin{tabular}{l|c|l|c|c}
			\hline
			%\tabincell{l}{\textit{Stage} of our \\embedding module} & \tabincell{c}{Top-1\\(\%)} &  \tabincell{c}{Params\\(M)} & \tabincell{c}{TFLOPs\\(G)} & \tabincell{c}{Speed\\(Hz)} \\
			\tabincell{l}{Pooling\\method } & Formula  &  \tabincell{l}{Matrix\\property} & \tabincell{l}{Structured\\Norm} & \tabincell{c}{Multi-\\head}  \\
			\hline
			GAP& $\frac{1}{q}\mathbf{Z}\mathbf{1}$ & \tabincell{l}{Vector}  & N/A & \xmark \\
			%\hline
			GCP& $\mathrm{MPN}\left(\frac{1}{q}\mathbf{Z}\mathbf{Z}^{T}\right)$ & \tabincell{l}{SPD}  & MPN & \xmark \\
			%\hdashline
			
			\tabincell{c}{MGCrP\\(ours)} & \tabincell{c}{$\left[\mathrm{svPN}\left(\frac{1}{q}\mathbf{W}_{1}\mathbf{Z}\mathbf{Z}^{T}\mathbf{R}_{1}^{T}\right),\ldots,\right.$ \\
				$ \left.\mathrm{svPN}\left(\frac{1}{q}\mathbf{W}_{h}\mathbf{Z}\mathbf{Z}^{T}\mathbf{R}_{h}^{T}\right)\right]\vspace{2pt}$
			}& \tabincell{l}{Asym} & $\mathrm{svPN}$ &\cmark\\
			%\hline
			%Attn & \tabincell{c}{$\mathrm{softmax}\big(\frac{1}{\sqrt{q}}\mathbf{Z}^{T}\mathbf{W}_{1}^{T}$\\$\times\mathbf{W}_{2}\mathbf{Z}\big)\mathbf{W}_{3}\mathbf{Z}$} &  \tabincell{l}{Asymmetric\\or non-square}  & LN&\cmark\\
			\hline
			
		\end{tabular}%
	\end{minipage}
	\caption{Differences of our MGCrP from the classical pooling methods. GAP produces  first-order, vectorial representations; GCP produces symmetric positive definite (SPD) matrices  to which matrix power normalization (MPN) is applicable;  MGCrP yields asymmetric  matrices, normalized by the proposed singular  value power normalization ($\mathrm{svPN}$). }\label{tab:cross-cov vs cov and attention}%
\end{table}

In the following, we first introduce cross-covariance pooling with single head. Given  an input matrix $\mathbf{Z}$, we perform two separate linear transformations, obtaining $\mathbf{X}=\mathbf{W}\mathbf{Z}$ and $\mathbf{Y=RZ}$, where $\mathbf{W}\in \mathbb{R}^{m\times p}$ and $\mathbf{R}\in \mathbb{R}^{n\times p}$ are learnable weight matrices. Then we compute cross-covariance matrix $\mathbf{Q}=\frac{1}{q}\mathbf{X}\mathbf{Y}^{T}$.   Previous works have shown that structured normalization plays an important role for GCP. However, as  $\mathbf{Q}$  is asymmetric (square or non-square),  existing normalization method for SPD matrices~\cite{PAMI_2020_MPN-COV} is not applicable. Thus we propose a new structured normalization method, called singular value power normalization (i.e., $\mathrm{svPN}$),  details of which are deferred to next section. Now we can define our global cross-covariance pooling: 
\begin{align}
\mathrm{GCrP}(\mathbf{Z})=\mathrm{svPN}\Big(\frac{1}{q}\mathbf{W}\mathbf{Z}\mathbf{Z}^{T}\mathbf{R}^{T}\Big)
\end{align} 
We continue to equip our  cross-covariance pooling with multi-head structure: 
\begin{align}
\mathrm{MGCrP}(\mathbf{Z})&=\Big[\mathrm{GCrP}_{1}(\mathbf{Z}),\cdots,\mathrm{GCrP}_{h}(\mathbf{Z})\Big]\nonumber\\
\mathrm{where}\ \mathrm{GCrP}_{i}(\mathbf{Z})&=\mathrm{svPN}\Big(\frac{1}{q}\mathbf{W}_{i}\mathbf{Z}\mathbf{Z}^{T}\mathbf{R}_{i}^{T}\Big)
\end{align}
Here $[\cdot]$ denotes concatenation operation; $\mathrm{GCrP}_{i}(\mathbf{Z})$ denotes the cross-covariance matrix of the $i$-th head, and $\mathbf{W}_{i}$ and $\mathbf{R}_{i}$ are two learnable linear projections.

Differences among GAP, GCP and MGCrP are summarized in Tab.~\ref{tab:cross-cov vs cov and attention}, where  $\mathbf{1}$ in the formula of GAP denotes a $q$-dimensional vector each component of which is 1, and we use state-of-the-art MPN~\cite{LiXWZ17} for GCP. Performance comparison among them is presented in Tab.~\ref{tab:comparison of fusion method}. Ablation analysis on MGCrP is given in  Sec.~\hyperref[subsection: ablation of  2nd-order pooling]{\textcolor{red}{4.2.2}}.

\subsection*{3.3.~Normalization of  Cross-covariance Matrix}\label{section:svPN}

We propose singular value power normalization (svPN)  for cross-covariance matrices. As svPN is computationally expensive, we further develop  a fast approximate   algorithm. 

\subsubsection*{3.3.1~~~~Singular Value Power Normalization} \label{subsection:svPN}

Our method  is motivated by MPN~\cite{PAMI_2020_MPN-COV}, an effective method for normalizing any covariance matrix $\mathbf{P}=\mathbf{Z}\mathbf{Z}^{T}$~\footnote{Without loss of generality, in Sec.~\hyperref[section:svPN]{\textcolor{red}{3.3}}, we omit the constant $1/q$ in representing a covariance  matrix or cross-covariance matrix for simplicity. } that is SPD. This normalization  consists in computation of  power of eigenvalues aligned with eigenvectors of $\mathbf{P}$.  In terms of  principal component analysis~\cite[Chap. 12]{Book-PRML},   the eigenvalues  of $\mathbf{P}$ in decreasing order amounts to from maximum to minimum variances of $\mathbf{Z}$ , while the corresponding eigenvectors of $\mathbf{P}$ characterize the principal components. Therefore,  MPN can be  interpreted as \textit{shrinking these variances  aligned with  the principal components}. 

Let us consider $\mathbf{u}^{T}\mathbf{X}\mathbf{Y}^{T}\mathbf{v}$ where $\mathbf{u}\in \mathbb{R}^{m}, \mathbf{v}\in \mathbb{R}^{n}$,  which indicates the cross-covariance between  projections of $\mathbf{X}$ on $\mathbf{u}$ and those of $\mathbf{Y}$ on $\mathbf{v}$. For simplification, we denote $\mathbf{Q}=\mathbf{X}\mathbf{Y}^{T}$. Actually, we have a  proposition about such cross-covariances. 

\begin{proposition}\label{proposition1-main}   
\textit{
	Given $\mathbf{u}_{i}$ and $\mathbf{v}_{i}$ where $i=1,\ldots, k-1$,  let us consider  the objective 
	\begin{align}\label{equ:kth-cross-cov-main}
	\max_{\|\mathbf{u}\|=\|\mathbf{v}\|=1} \;&\mathbf{u}^{T}\mathbf{X}\mathbf{Y}^{T}\mathbf{v}\\
	\mathrm{s.t.}\;\;&\mathbf{u}_{i}^{T}\mathbf{u}=0, \;\mathbf{v}_{i}^{T}\mathbf{v}=0, \;i<k.\nonumber
	\end{align} 
	By inductively optimizing~(\ref{equ:kth-cross-cov-main}) for $k=1,\ldots, \min(m,n)$, we can obtain,  in decreasing order,  the $k$-th largest cross-covariance $\mathbf{u}_{k}^{T}\mathbf{X}\mathbf{Y}^{T}\mathbf{v}_{k}$, which  is equal to the $k$-th singular value  $\lambda_{k}$ of $\mathbf{Q}$ while $\mathbf{u}_{k}$ and $\mathbf{v}_{k}$ are the corresponding  left and right singular vectors, respectively.
	}
\end{proposition}
\vspace{-2pt}In light of this proposition,  we  can define our normalization by \textit{shrinking the cross-covariances between $\mathbf{X}$ and $\mathbf{Y}$ aligned with the left and right singular vectors of $\mathbf{Q}$}: 
\begin{align}\label{equ:svPN_alpha_SVD}
\mathrm{svPN}(\mathbf{Q})=\sum_{i=1}^{\min(m,n)}\lambda_{i}^{\alpha}\mathbf{u}_{i}\mathbf{v}_{i}^{T},
\end{align} 
where $0\!\!<\!\!\alpha\!\!<\!\!1$.  Our $\mathrm{svPN}$ can be performed accurately via SVD, which, 
however,  is computationally  expensive, as the SVD algorithm is GPU-unfriendly~\cite{LiXWG18}. We give a proof of Proposition~\ref{proposition1-main} and the backpropagation  of svPN via SVD in the supplement~\ref{suppsection:Proof of Proposition} and~\ref{suppsection:Backpropagation}, respectively. 

\subsubsection*{3.3.2~~~~Fast Approximate Normalization Algorithm}\label{subsection:approximate svPN}

Based on low-rank assumption widely used in machine learning~\cite{low-rank-workshop}, we can efficiently implement  approximate  normalization by only estimating few largest singular  values.
We use an iterative method  to consecutively estimate the  singular values. Given an initial vector $\mathbf{v}^{(0)}$, the iterative  procedure takes the following  form~\cite{SVD_TPAMI_1982}: 
\begin{align}\label{equ:iteration_u_and_v}
\mathbf{u}^{(j+1)}=\dfrac{\mathbf{Q}\mathbf{v}^{(j)}}{\|\mathbf{Q}\mathbf{v}^{(j)}\|},\;
\mathbf{v}^{(j+1)}=\dfrac{\mathbf{Q}^{T}\mathbf{u}^{(j+1)}}{\|\mathbf{Q}^{T}\mathbf{u}^{(j+1)}\|}
\end{align}
where the superscript $j$ denotes the $j$-th iteration. After a few iterations, we obtain approximately the largest singular value  $\hat{\lambda}_{1}=\|\mathbf{Q}^{T}\mathbf{u}^{(j+1)}\|$ and the corresponding left singular vector  $\hat{\mathbf{u}}_{1}=\mathbf{u}^{(j+1)}$ and right one   $\hat{\mathbf{v}}_{1}=\mathbf{v}^{(j+1)}$. 

Suppose we have the $k$-th ($k\geq 1$) largest singular value, we deflate matrix $\mathbf{Q}$ to obtain
\begin{align}\label{equ:deflation}
\mathbf{Q}'=\mathbf{Q}-\sum_{i=1}^{k}\hat{\lambda}_{i}\hat{\mathbf{u}}_{i}\hat{\mathbf{v}}_{i}^{T}
\end{align}
For $\mathbf{Q}'$, we can perform iteration with  Eq.~(\ref{equ:iteration_u_and_v}) to achieve approximately the $(k+1)$-th largest singular value $\hat{\lambda}_{k+1}$ and the corresponding singular vectors $\hat{\mathbf{u}}_{k+1}$ and $\hat{\mathbf{v}}_{k+1}$. 
The deflation~(\ref{equ:deflation})  and the iteration~(\ref{equ:iteration_u_and_v}) can be repeated.  Given $r$ largest singular values, we define  the approximate  normalization as
\begin{align}\label{equ:approximated svPN_alpha_SVD}
\widehat{\mathrm{sv}}\mathrm{PN}(\mathbf{Q})\!=\!\sum_{i=1}^{r-1}\hat{\lambda}_{i}^{\alpha}\hat{\mathbf{u}}_{i}\hat{\mathbf{v}}_{i}^{T}
\!+\!\dfrac{1}{\hat{\lambda}_{r}^{1-\alpha}}\big(\mathbf{Q}\!-\!\sum_{i=1}^{r-1}\hat{\lambda}_{i}\hat{\mathbf{u}}_{i}\hat{\mathbf{v}}_{i}^{T}\big)
\end{align} 
It shrinks the 1st to the $(r-1)$-th singular values aligned with the corresponding singular vectors, while shrinking the remaining ones with  the $r$-th largest singular value. Note that $\widehat{\mathrm{sv}}\mathrm{PN}(\mathbf{Q})$ reduces to $\mathbf{Q}/\hat{\lambda}_{1}^{1-\alpha}$ if we only use the largest singular value. 

%-------------------------------------------------------------------------
\section*{4.~Experiments}\label{section:experiments}

We first introduce  experimental setting in Sec.~\hyperref[subsection:setting]{\textcolor{red}{4.1}}. Then we evaluate the proposed  methods for  computer vision (CV) tasks   and  natural language processing (NLP) tasks  in Sec.~\hyperref[subsection: image classification]{\textcolor{red}{4.2}} and Sec.~\hyperref[subsection:text classification]{\textcolor{red}{4.3}}, respectively. We train models with 8 NVIDIA 2080Ti GPUs based on PyTorch framework. Our code will be open-sourced after acceptance.

\subsection*{4.1.~Experimental Setting}\label{subsection:setting}

Here we briefly describe the benchmarks and training strategy for both CV and NLP tasks. Details on benchmark statistics, task description and  hyper-parameters setting are given in the supplement~\ref{suppsection:detailed}. 

\vspace{4pt}\noindent\textbf{Datasets}
For CV tasks, we evaluate on ILSVRC  ImageNet benchmark~\cite{ILSVRC15,imagenet_cvpr09}, which has 1.28M images for training and 50K images for test.  Furthermore,  we adopt a  more challenging dataset (i.e.,   ImageNet-A~\cite{ImageNet-A}) for evaluation,  which consists of real-world adversarial images, involving  heterogeneous and varied distribution shift.  

For NLP tasks, following the common practice~\cite{GPT-1,DBLP:conf/naacl/DevlinCLT19},  we fine-tune the transformer models pre-trained in an unsupervised manner on large-scale corpus on four downstream tasks, i.e., Corpus of Linguistic Acceptability (CoLA)~\cite{warstadt2018neural}, Recognizing Textural Entailment (RTE)~\cite{bentivogli2009fifth},  Multi-Genre Natural Language Inference (MNLI)~\cite{williams2018broad} and  Stanford Question Answering (QNLI)~\cite{rajpurkar2016squad}.

\vspace{4pt}\noindent\textbf{Training  strategy}
In image classification, we train our SoT models on ImageNet  from scratch. Besides scale, color and flip  augmentations~\cite{Simonyan15, He_2016_CVPR}, following~\cite{T2T_ICCV21}, we also adopt  mixup~\cite{zhang2018mixup}, randAugment~\cite{cubuk2020randaugment}, cutmix~\cite{yun2019cutmix}, and  label smoothing~\cite{Szegedy_2015_CVPR}.  We use AdamW~\cite{loshchilov2018decoupled} algorithm with warmup for network optimization and  cosine annealing schedule for learning rate.
Detailed training strategies about optimizers, hyper-parameters, etc.,  are given in the supplement~\ref{suppsection:detailed}.  Also, we present in detail  the hyper-parameter setting  on  natural language processing tasks in the supplement~\ref{suppsection:detailed}.

\subsection*{4.2.~Image Classification for CV Tasks}\label{subsection: image classification}

To make our extensive evaluations computationally feasible on ImageNet, in Sec.~\hyperref[subsection: experiments-how novel]{\textcolor{red}{4.2.1}} and Sec.~\hyperref[subsection: ablation of  2nd-order pooling]{\textcolor{red}{4.2.2}}, we use the 7-layer SoT model; besides, we re-scale the images such that their short sizes are 128 and 112$\times$112 patches are cropped as network inputs. In Sec.~\hyperref[subsection:experiment-comparion with state-of-the-art]{\textcolor{red}{4.2.3}},  we compare with state-of-the-art methods  using standard protocol on ImageNet.

\vspace{-6pt}\subsubsection*{4.2.1.~~~How Does Our Classification Head Perform?} \label{subsection: experiments-how novel}

We first establish a strong baseline  model that only uses the classification token. Based on this strong baseline, we compare different fusion schemes and pooling methods.

\vspace{6pt}\noindent\textbf{Baseline based on conventional classification head}   
Several works~\cite{T2T_ICCV21,PSViT_ICCV21} improve the simple embedding method of ViT~\cite{ViT} (i.e., a naive linear projection of fixed patches). Specifically, T2T~\cite{T2T_ICCV21} introduces soft-split operations called Tokens-to-token and PS-ViT~\cite{PSViT_ICCV21} proposes a progressively sampling  strategy built upon a convolution stem to learn the local structure of images. In contrast, we design a small, hierarchical module based on off-the-shelf convolutions for token embedding, and evaluate various types of convolution blocks. Tab.~\ref{tab:our baseline model} compares these baseline models  where only classification token is used for the classifier. It can be seen that both  T2T and PS-ViT clearly improve over ViT, while all of our models  perform much better than the two variants while having comparable parameters and FLOPs. Our embedding module with DenseNet block obtains the best result (73.13\%), establishing a strong baseline. This token embedding module will be used across our family of transformer models.

\begin{table}[t]
	\centering
	\footnotesize
	\setlength{\tabcolsep}{2pt}
	\renewcommand\arraystretch{1.1}
	
	\begin{minipage}{1\linewidth}
		\begin{subtable}{1\linewidth}
			\centering
			\begin{tabular}{c|l|c|c|c}
				\hline
				%\tabincell{l}{\textit{Stage} of our \\embedding module} & \tabincell{c}{Top-1\\(\%)} &  \tabincell{c}{Params\\(M)} & \tabincell{c}{TFLOPs\\(G)} & \tabincell{c}{Speed\\(Hz)} \\
				Model &\tabincell{c}{Token embedding} & \tabincell{c}{Top-1\\ (\%)} &  \tabincell{c}{Params\\ (M)} & \tabincell{c}{FLOPs\\ (G)}  \\
				\hline
				\multirow{7}{*}{\tabincell{l}{Baseline\\(classification\\ token only)}}&Naive linear proj~\cite{ViT} & 66.25 & 3.98 & 0.73  \\
				&Tokens-to-token~\cite{T2T_ICCV21} & 68.30 & 4.25 & 0.81  \\
				&Progressive sampling~\cite{PSViT_ICCV21} & 70.48 & 4.13 & 0.90 \\
				%\cdashline{2-5}
				\cline{2-5}
				&ResNet block (ours)&  72.51 &4.28 & 0.88  \\
				&Inception block (ours) & 71.61 &4.09&  0.81 \\
				&DenseNet block (ours)& \textbf{73.13} & 4.23  & 1.06 \\
				&Non-local block (ours)& 71.45  & 4.17  &0.89\\
				\hline				
			\end{tabular}%
			\caption{Results of baselines using conventional classification head.}\label{tab:our baseline model}%
			
		\end{subtable}
	\end{minipage}

\vspace{6pt}
	
	\begin{minipage}{1\linewidth}
	\begin{subtable}{1\linewidth}
		\centering
		\begin{tabular}{c|c|c|c|c}
			\hline
			\tabincell{c}{Fusion scheme}  & Pool method & Repr. size & Top-1 (\%) & Params (M) \\
			\hline
			\multicolumn{2}{c|}{Baseline} & 256 &  73.13 &  4.23\\
			%\cline{4-7}
			\multicolumn{2}{c|}{(classification token only)} &  1280 & 73.43 & 5.57\\
			\hline
			\multirow{3}{*}{$\mathrm{aggr\_all}$} & GAP & 256 &73.56 & \textbf{4.22} \\
			&  GCP   & 1176 & 74.12  &    5.15  \\
			&  MGCrP & 1176 & 74.74  & 5.20 \\
			\hline																 
			\multirow{3}{*}{$\mathrm{concat}$} & GAP & 512 &73.84&4.73 \\
			& GCP   & 1432 & 75.37 & 5.66   \\
			& MGCrP & 1432 & 75.85 &   5.71 \\
			\hline
			\multirow{3}{*}{$\mathrm{sum}$} & GAP & 256 & 73.85  &4.47\\
			&  GCP   & 1176 & 75.23   & 5.40\\
			&  MGCrP & 1176 &\textbf{75.97} &5.44\\
			\hline
			\multirow{3}{*}{$\mathrm{late}$}& GAP  & 256 & 73.27 & 4.47\\
			&  GCP   & 1176 &73.46&5.40\\
			&  MGCrP & 1176 &73.98&5.46\\
			\hline	
		\end{tabular}%
		\caption{Results using our classification head. }\label{tab:comparison of fusion method}%
	\end{subtable}
\end{minipage}
	\caption{Evaluation of proposed classification head. (a) We build a strong baseline which only uses the  classification token. (b) We compare different fusion schemes along with pooling methods, based on the strong baseline. }
	\label{table: effect of fusion}	
	
\end{table}

\begin{table*}
	\centering
	\begin{subtable}[t]{0.38\linewidth}
		\centering
		\setlength{\tabcolsep}{2pt}
		\renewcommand\arraystretch{1.3}
		\footnotesize
		\vspace{-43pt}
		\begin{tabular}{c|c|c|c|c|c}
			\hline
			$h$ & 1 & 2 & 4 & 6 & 8 \\
			\hline
			($m$,$n$) & (32,32) & (24,24) & (16,16)  & (14,14)  & (12,12) \\
			\hline
			%dimension & 1k   & 1.1k  & 1k & 1.1k & 1.1k\\
			%\hline
			Top-1 (\%) & 75.14 & 75.36 & 75.24 & \textbf{75.97} & 75.37\\
			\hline				
		\end{tabular}%
		\caption{Effect of  head number  $h$ given fixed Repr. size ($\sim$$\mathrm{1K}$). }\label{tab:comparison of_head}%	
		\centering
		% 		\setlength{\tabcolsep}{3pt}
		% 	\renewcommand\arraystretch{1.2}
		% 		\footnotesize
		\vspace{5pt}
		\begin{tabular}{c|c|c|c|c|c}
			\hline
			Repr. size & 0.5K & 1K  & 2K & 3K & 6K \\
			\hline
			($m$,$n$) & (9,9) & (14,14) & (18, 18) & (24,24) & (32,32)  \\
			\hline
			Top-1 (\%) & 74.38& 75.97 & 76.24 & 76.73&\textbf{77.01}\\
			\hline				
		\end{tabular}%
		\caption{Effect of Repr. size given fixed head number ($h=6$). }\label{tab:comparison of_dimension}%				
	\end{subtable}
	\begin{subtable}[t]{0.33\linewidth}
		\centering
		\setlength{\tabcolsep}{2pt}
		\footnotesize
		\vspace{-15mm}
		\begin{tabular}{c|c|c|c|c}					
			\hline
			Method & \multicolumn{2}{c|}{Setting} & \parbox{8mm}{\centering Top-1\\ (\%)} & \tabincell{c}{Speed\\ (Hz)}\\
			\hline
			\multirow{3}{*}{$\mathrm{svPN}$} & \multirow{3}{*}{$\alpha$} &  0.3 & 75.74  &\multirow{3}{*}{110} \\
			\cline{3-4}
			&   & 0.5 & \textbf{76.13} & \\
			\cline{3-4}
			&   & 0.7 & 75.45 & \\
			\hline
			\multirow{5}{*}{$\widehat{\mathrm{sv}}\mathrm{PN}$} & \multirow{5}{*}{\tabincell{c}{(\#sv,\#iter)\\ $\alpha$=0.5}} & $(1,1)$ & 75.97 & \textbf{2226}\\
			\cline{3-5}
			&   & $(1,3)$ & 75.82 & 2207\\
			\cline{3-5}
			&   & $(1,5)$ & 74.11 &2188\\
			\cline{3-5}
			&   & $(2,1)$ & 73.51 &2207 \\
			\cline{3-5}
			&   & $(3,1)$ & 74.89 & 2188\\
			\hline
		\end{tabular}%
		\caption{Exact normalization  versus approximate one.}\label{tab: eact vs approximate svPN}%
	\end{subtable}
	\begin{subtable}[t]{0.25\linewidth}
		\centering
		\setlength{\tabcolsep}{4pt}
		\renewcommand\arraystretch{1.2}
		\footnotesize		
		%\vspace{1mm}
				\begin{tabular}{r|l|c}
			\hline
			Method  &  Top-1 (\%) & Speed (Hz)\\
			\hline
			--   &74.82 & \textbf{2248}\\
			$1/\tau$  &75.13$_{\,\text{0.31}\uparrow}$ & 2226\\
			EPN~\cite{lin2015bilinear}   &73.29$_{\,\text{1.53}\downarrow}$  &2206\\
			LN~\cite{Layer_Norm_arXiv_2016}  &75.25$_{\,\text{0.43}\uparrow}$ & 2245\\
			\hline
			$\widehat{\mathrm{sv}}\mathrm{PN}$ & 75.97$_{\,\text{1.15}\uparrow}$  &2226\\
			$\mathrm{sv}\mathrm{PN}$  & \textbf{76.13}$_{\,\textbf{\text{1.31}}\uparrow}$  &110\\
			\hline
		\end{tabular}%
		\caption{Comparison with different normalization methods.}\label{tab:comparison of normalization}%			
	\end{subtable}
	\caption{Ablation analysis of MGCrP and normalization.}
\end{table*}

\vspace{6pt}\noindent\textbf{Effect of our classification head}  \label{subsection: experiment-fusion}
Tab.~\ref{tab:comparison of fusion method} evaluates the proposed classification head on the basis of the strong baseline. We adopt iterative square-root normalization (iSQRT)~\cite{LiXWG18} for GCP which is a fast version of matrix power normalization (MPN)~\cite{LiXWZ17}. For our MGCrP, we use $\widehat{\mathrm{sv}}\mathrm{PN}$ with the single largest singular value. According to the results in Tab.~\ref{tab:comparison of fusion method},   we have two observations. (1) Fundamentally, all fusion schemes improve the baseline whatever the pooling function is, which indicates that explicit combination of the word tokens indeed benefits the transformer models.  Among the fusion methods, the $\mathrm{aggr\_all}$ scheme is superior to the $\mathrm{late}$ scheme  which only slightly improves the baseline, while both of the  $\mathrm{sum}$  scheme and  $\mathrm{concat}$ scheme  perform much better than the other two fusion methods.  (2)  The second-order pooling   outperforms the first-order pooling by a large margin regardless of fusion method, which is consistent with previous conclusion drawn under the CNN architectures~\cite{pami/LinRM18,PAMI_2020_MPN-COV}. For any fusion method, our MGCrP performs better than GCP by about 0.5\%, suggesting that our multi-headed cross-covariance pooling has more powerful representation capability. We note that  the $\mathrm{sum}$ scheme with MGCrP obtains the highest accuracy of $75.97\%$. 

The second-order pooling enlarges  the representation size (Repr. size), leading to more parameters than the baseline. For fair comparison, for the baseline model,  we add a linear projection layer after the classification token, increasing its dimension to 1280; as a result, its accuracy  increases,  but only slightly (0.3\%, 2nd row in Tab.~\ref{tab:comparison of fusion method}), which is still much lower than the fusion method. This suggests that the performance increase of the fusion methods is mainly attributed to more powerful representation capability rather than capacity growth. Note that all our fusion methods bring negligible increase of FLOPs, compared to the baseline.

\subsubsection*{4.2.2.~~~Ablation Study of MGCrP and Normalization}\label{subsection: ablation of  2nd-order pooling}

For the MGCrP module, we first evaluate the number of heads and representation size.  After that, we assess the exact  normalization (i.e., $\mathrm{svPN}$) against the approximate one (i.e., $\widehat{\mathrm{sv}}\mathrm{PN}$). Finally, we compare with  other normalization methods for cross-covariance matrices. 

\vspace{3pt}\noindent\textbf{Number of heads and representation size}   The representation size (Repr. size) of MGCrP is equal to $h\times m \times n$, where $h$, $m$ and $n$ are  number of heads, and dimensions  of  two linear projections, respectively. Exhaustive grid search for these hyper-parameters of MGCrP is computationally prohibitive. For simplification,  we set $m=n$, and search for  optimal $h$ by fixing the Repr. size, followed by evaluation of  the Repr. size with fixed number of heads $h$ just determined.  Tab.~\ref{tab:comparison of_head} shows  accuracy versus $h$ when the Repr. size is fixed to about 1K. It can be seen that $h=6$ achieves the best result. By setting $h$ to 6, Tab.~\ref{tab:comparison of_dimension} shows  the effect of the representation size on  performance. We can see that the accuracy consistently increases as the Repr. size grows while less than 3K;  however,  further doubling  Repr. size to 6K brings minor increase of  accuracy, which suggests that the performance tends to saturate. We use six heads for MGCrP across the paper, unless otherwise specified.

\vspace{3pt}\noindent\textbf{Exact normalization against approximate one}   Tab.~\ref{tab: eact vs approximate svPN} (upper panel) shows  the effect of exponent $\alpha$  (Eq.~\ref{equ:svPN_alpha_SVD}) for the exact normalization $\mathrm{svPN}$,  where $\alpha=0.5$ achieves the highest accuracy. However, svPN via SVD is computationally very expensive, running only at 110 Hz. By setting $\alpha$ to 0.5,  we demonstrate, in Tab.~\ref{tab: eact vs approximate svPN} (lower panel), the effect of the number of singular values ($\#\mathrm{sv}$) and the number of iterations ($\#\mathrm{iter}$) on $\widehat{\mathrm{sv}}\mathrm{PN}$ (Eq.~\ref{equ:approximated svPN_alpha_SVD}). We note that the  approximate normalization  is slightly inferior to but runs 20 times faster than its exact counterpart.  With only the  largest singular  value, increase of iteration number brings no gains;  If we use two or three largest singular values, we observe performance decline. We conjecture   the reason is that  more iterations accumulate numerical errors, leading to performance decline. Notably, $\widehat{\mathrm{sv}}\mathrm{PN}$ with the single largest eigenvalue and one iteration achieves a very competitive accuracy of 75.97\% with the fastest speed of 2226 Hz, and this setting is used throughout, unless otherwise specified.

\vspace{3pt}\noindent\textbf{Different normalization methods} Besides the proposed normalization method, there are several other options to  normalize the cross-covariance matrices, including layer normalization (LN)~\cite{Layer_Norm_arXiv_2016},  EPN~\cite{lin2015bilinear} and an adaptive scaling. EPN consists in signed square root for each element followed by $\ell_{2}$ normalization. In contrast to $\widehat{\mathrm{sv}}\mathrm{PN}$ which  scales the cross-covariance matrix by $1/\lambda_{1}^{1-\alpha}$, we design the adaptive scaling which learns a scalar $1/\tau>0$  to calibrate the cross-covariance matrix.    The comparison results are given in Tab.~\ref{tab:comparison of normalization}.   We can see that all normalization methods except EPN improve over the baseline that has no normalization. In particular, our normalization methods perform much better than the competitors, and improve the baseline by~$\sim$~1.2\%, suggesting  superiority of our normalization method for  cross-covariance matrices.

\begin{table}[htb!]
	\centering
	\setlength{\tabcolsep}{2pt}
	\footnotesize
	\renewcommand\arraystretch{1.0}
	\begin{tabular}{c}
		\begin{minipage}{1\linewidth}
			\begin{subtable}{1\linewidth}
				\centering
				\footnotesize
				\setlength{\tabcolsep}{3pt}
				\begin{tabular}{l|c|c|c|c}
					\hline
					\multirow{2}{*}{Model}  & Params & FLOPs & ImageNet & ImageNet-A    \\
					& (M) & (G) & Top-1 (\%)  & Top-1 (\%)  \\
					\hline
					DeiT-T~\cite{DeiT_ICML} & 5.7  & 1.3 & 72.2  & 7.3 \\
					T2T-ViT-7~\cite{T2T_ICCV21} & 4.3  & 1.2 & 71.7  &  6.1  \\
					T2T-ViT-12~\cite{T2T_ICCV21} & 6.9  & 2.2 & 76.5  &  12.2  \\
					PS-ViT-Ti/14~\cite{PSViT_ICCV21} & 4.8  & 1.6 & 75.6  & --  \\
					PVT-T~\cite{PVT_ICCV21} & 13.2  & 1.9 & 75.1  & 7.9  \\
					PiT-Ti~\cite{PiT_ICCV21} & 4.9  & 0.7 & 73.0  & 6.2   \\
					iRPE-K-T~\cite{iRPE_ICCV21} & 6.0  & 1.3 & 73.7  &  8.8 \\
					AutoFormer-tiny~\cite{AutoFormer_ICCV21} & 5.7 & 1.3 & 74.7 &  10.3 \\
					%\hdashline
					SoT-Tiny (ours) & 7.7 & 2.5 & \textbf{80.3} & \textbf{21.5} \\
					\hline 
					DeiT-T+ours & 7.0 & 2.3 &  78.6$_{6.4\uparrow}$ & 17.5$_{10.2\uparrow}$ \\
					iRPE-K-T+ours &  7.0 &2.3  & 79.0$_{5.3\uparrow}$  & 18.2$_{\;\;9.4\uparrow}$  \\
					T2T-ViT-12+ours & 6.9 & 2.3 & 79.4$_{2.9\uparrow}$ & 15.4$_{\;\;3.2\uparrow}$   \\
					\hline
					
				\end{tabular}%
				\setlength{\abovecaptionskip}{1.5pt}
				\setlength{\belowcaptionskip}{4pt}
				\caption{Comparison with light-weight models.}\label{subtab:small}
				
			\end{subtable}%\\
		\end{minipage} \\
		
		\begin{minipage}{1\linewidth}
			\begin{subtable}{1\linewidth}
				\centering
				\footnotesize
				\setlength{\tabcolsep}{3pt}
				\begin{tabular}{l|c|c|c|c}
					\hline
					\multirow{2}{*}{Model}  & Params & FLOPs & ImageNet & ImageNet-A    \\
					& (M) & (G) & Top-1 (\%)  & Top-1 (\%)    \\
					\hline
					DeiT-S~\cite{DeiT_ICML} & 22.1  & 4.6 & 79.8  & 18.9   \\
					T2T-ViT-14~\cite{T2T_ICCV21} & 21.5  & 5.2 & 81.5  & 23.9  \\
					PVT-S~\cite{PVT_ICCV21} & 24.5  & 3.8 & 79.8  & 18.0  \\
					PS-ViT-B/10~\cite{PSViT_ICCV21} & 21.3  & 3.1 & 80.6  & -- \\
					PS-ViT-B/14~\cite{PSViT_ICCV21} & 21.3  & 5.4 & 81.7  &  27.3   \\
					PS-ViT-B/18~\cite{PSViT_ICCV21} & 21.3  & 8.8 & 82.3  &  31.7  \\
					iRPE-QKV-S~\cite{iRPE_ICCV21} & 22.0  & 4.9 & 81.4  & 25.0   \\
					PiT-S~\cite{PiT_ICCV21} & 23.5  & 2.9 & 80.9  & 21.7  \\
					AutoFormer-small~\cite{AutoFormer_ICCV21} & 22.9 & 5.1 & 81.7 & 25.7 \\
					Conformer-Ti~\cite{Conformer_ICCV21} & 23.5  & 5.2 & 81.3  & 27.2  \\
					Swin-T~\cite{Swin_ICCV21} & 28.3  & 4.5 & 81.3  & 21.6 \\
					%\hdashline
					SoT-Small (ours) & 26.9 & 5.8 & \textbf{82.7} & \textbf{31.8}  \\
					\hline
					DeiT-S+ours & 25.6 & 5.5 & 82.7$_{2.9\uparrow}$ & 32.2$_{13.3\uparrow}$\\
					T2T-ViT-14+ours & 24.4 & 5.4 & 82.6$_{1.1\uparrow}$ & 27.1$_{\;\;3.2\uparrow}$  \\
					Swin-T+ours & 31.6 & 6.0 &	83.0$_{1.7\uparrow}$&  33.5$_{11.9\uparrow}$  \\
					Conformer-Ti+ours & 30.6& 6.3&83.0$_{1.7\uparrow}$ &36.4$_{\;\;9.2\uparrow}$  \\
					\hline
					
				\end{tabular}%
				\setlength{\abovecaptionskip}{1.5pt}
				\setlength{\belowcaptionskip}{4pt}
				\caption{Comparison with  middle-sized models. }\label{subtab:middle}
			\end{subtable}%\\
		\end{minipage} \\
		
		\begin{minipage}{1\linewidth}
			\begin{subtable}{1\linewidth}
				\centering
				\footnotesize
				\setlength{\tabcolsep}{3pt}
				\renewcommand\arraystretch{1.2}
				\begin{tabular}{l|c|c|c|c}
					\hline
					\multirow{2}{*}{Model}  & Params & FLOPs & ImageNet & ImageNet-A    \\
					& (M) & (G) & Top-1 (\%)  & Top-1 (\%)    \\
					\hline
					DeiT-B~\cite{DeiT_ICML} & 86.6  & 17.6 & 81.8  & 27.4  \\
					T2T-ViT-24~\cite{T2T_ICCV21} & 64.1  & 15.0 & 82.3  & 28.9   \\
					PVT-L~\cite{PVT_ICCV21} & 61.4  & 9.8 & 81.7  & 26.6   \\
					iRPE-K-B~\cite{iRPE_ICCV21} & 87.0  & 17.7 & 82.4  & 31.8  \\
					PiT-B~\cite{PiT_ICCV21} & 73.8  & 12.5 & 82.0  & 33.9   \\
					AutoFormer-base~\cite{AutoFormer_ICCV21} & 54.0 & 11.0 & 82.4 &  28.8  \\
					Swin-B~\cite{Swin_ICCV21} & 87.8  & 15.4 & \textbf{83.5}  & \textbf{35.8}   \\
					%\hdashline
					SoT-Base  (ours)& 76.8  & 14.5  &\textbf{83.5}  & 34.6  \\
					\hline 	 
					DeiT-B+ours & 94.9 & 18.2 & 82.9$_{1.1\uparrow}$ & 29.1$_{1.7\uparrow}$   \\
					T2T-ViT-24+ours & 72.1 & 15.5 & 83.3$_{1.0\uparrow}$ & 30.1$_{1.2\uparrow}$   \\
					Swin-B+ours & 95.9  & 16.9  & 84.0$_{0.5\uparrow}$ &  42.9$_{7.1\uparrow}$ \\
					\hline
					
				\end{tabular}%
				\setlength{\abovecaptionskip}{1.5pt}
				\caption{Comparison with heavyweight models. }\label{subtab:large}
			\end{subtable}%
		\end{minipage}\\
	\end{tabular}
	\caption{Comparison with state-of-the-art vision  transformer models on image classification tasks.}
	\label{tab:sota}

\end{table}

\subsubsection*{4.2.3.~~~Comparison with  State-of-the-art Methods}\label{subsection:experiment-comparion with state-of-the-art}

 We present comparisons with a series of vision  transformer models. Tabs.~\ref{subtab:small}, \ref{subtab:middle} and \ref{subtab:large} show  comparison results with  lightweight models, middle-sized models and  heavyweight models, respectively.  In light of  these results we can draw the following three conclusions. (1) As regards our family of transformer models, on ImageNet, SoT-Tiny significantly outperforms  the competing light-weight transformers by 3.8\%. When the networks deepens, the performance gaps between  our models and the competitors become smaller. This is natural as the network gets deeper, further performance increase becomes more difficult~\cite{He_2016_CVPR}. (2) When we equip state-of-the-art architectures with our method, on ImageNet, we can invariably observe consistent benefits while introducing small, additional cost. For light-weight models, the gains are  substantial, i.e., 6.4\%, 5.3\% and 2.9\% for DeiT-T, iRPE-K-T and T2T-ViT-12, respectively. For middle-sized models, the improvements are 1.1\%$\sim$2.9\%. Even for very strong  heavyweight models, we can still achieve gains of 0.5\%$\sim$1.0\%. Note that Swin-Transformer models have  no classification token, so we use our MGCrP in place of the original GAP. 
 These comparison results demonstrate that our method well generalizes to different vision  transformer architectures.   (3) On ImageNet-A,  our SoT-Tiny is superior across the light-weight models, while our SoT-Small and SoT-Base are very competitive compared to state-of-the-art models. Notably, equipped with our method, the compared state-of-the-art methods can achieve impressive improvements, i.e., 3.2\%$\sim$10.2\% for light-weight models, 3.2\%$\sim$13.3\% for middle-sized models and 1.2\%$\sim$7.1\% for heavyweight models. These results indicate that our classification head can substantially enhance robustness of different architectures.

\subsection*{4.3~~Text Classification for  NLP Tasks}\label{subsection:text classification}

At last, we evaluate our classification head on natural language processing tasks.  Note that our purpose here is not to achieve state-of-the-art performance; instead, we intend to show  how our classification head performs against the conventional classification head.  Two kinds of pre-trained transformer models are used, namely GPT~\cite{GPT-1} as well as  BERT~\cite{DBLP:conf/naacl/DevlinCLT19} and its stronger variants (i.e., SpanBERT~\cite{joshi2020spanbert} and RoBERTa~\cite{roberta}).  According to Tab.~\ref{tab:text_classification}, on CoLA and RTE,  our method with GPT models improves over the conventional one by  2.18\% or more, while with BERT models and the variants, we achieve 1.19\%$\sim$6.29\% gains in accuracy. As opposed to  CoLA and RTE, the improvement on MNLI and QNLI is  not that large: with GPT we achieve 0.31\% gains for GPT and   0.27\%$\sim$0.73\%  for BERT or its  variants. Note that MNLI and QNLI are much bigger than CoLA and RTE, both containing similar sentences with pre-training datasets; consequently, the performance of individual models may tend to saturate and further improvement becomes difficult.  Notably, the magnitude of performance boost by using our method preserves  for stronger models, e.g., the gains over stronger RoBERTa are comparable to those over BERT. It is well-known that real-world tasks often have  limited labeled data since human annotations are  expensive and laborious. For such tasks,  our method is more preferred as it can provide nontrivial  performance increase over  the conventional method.

\begin{table}[t]
	\centering
	\footnotesize
	\setlength{\tabcolsep}{2pt}
	\renewcommand\arraystretch{1.0}
	\begin{tabular}{l|l|l|l|l}%{lllll}
		\hline
		Model & CoLA &  RTE & MNLI & QNLI\\
		\hline
		GPT~\cite{GPT-1}        & 54.32 & 63.17 &82.10&86.36\\
		GPT+ours & 57.25$_{2.93\uparrow}$ & 65.35$_{2.18\uparrow}$ &82.41$_{0.31\uparrow}$ &87.13$_{0.77\uparrow}$\\
		\hline
		\hline
		BERT-base~\cite{DBLP:conf/naacl/DevlinCLT19} & 54.82 & 67.15 & 83.47 &90.11 \\
		BERT-base+ours & 58.03$_{3.21\uparrow}$ & 69.31$_{2.16\uparrow}$ & 84.20$_{0.73\uparrow}$ &90.78$_{0.67\uparrow}$\\
		BERT-large~\cite{DBLP:conf/naacl/DevlinCLT19} & 60.63  & 73.65  & 85.90 &91.82 \\
		BERT-large+ours &  61.82$_{1.19\uparrow}$ & 75.09$_{1.44\uparrow}$ & 86.46$_{0.56\uparrow}$ &92.37$_{0.55\uparrow}$\\
		\hline
		SpanBERT-base~\cite{joshi2020spanbert} & 57.48 &  73.65 & 85.53 &92.71\\
		SpanBERT-base+ours & 63.77$_{6.29\uparrow}$ & 77.26$_{3.61\uparrow}$  &  86.13$_{0.60\uparrow}$   &93.31$_{0.60\uparrow}$ \\
		SpanBERT-large~\cite{joshi2020spanbert} &  64.32 & 78.34 &87.89 &94.22 \\
		SpanBERT-large+ours &  65.94$_{1.62\uparrow}$ & 79.79$_{1.45\uparrow}$  &88.16$_{0.27\uparrow}$  & 94.49$_{0.27\uparrow}$\\
		\hline
		RoBERTa-base~\cite{roberta} & 61.58  & 77.60  & 87.50&92.70\\
		RoBERTa-base+ours &65.28$_{3.70\uparrow}$ & 80.50$_{2.90\uparrow}$  &87.90$_{0.40\uparrow}$  &93.10$_{0.40\uparrow}$    \\
		RoBERTa-large~\cite{roberta} &  67.98 &  86.60 & 90.20&94.70 \\
		RoBERTa-large+ours &  70.90$_{2.92\uparrow}$ & 88.10$_{1.50\uparrow}$  &90.50$_{0.30\uparrow}$  & 95.00$_{0.30\uparrow}$   \\
		\hline
		
	\end{tabular}%
	\caption{Performance improvement over language transformer models on text classification tasks. }\label{tab:text_classification}%
\end{table}

%------------------------------------------------------------------------
\section*{5.~Conclusion}

In this paper, we propose a novel second-order transformer (SoT) model.  The key of our SoT is  a novel classification  head which exploits simultaneously word tokens and classification token. It goes beyond, for the first time as far as we know, the prevalent classification paradigm of transformers which exclusively  use the classification  token. We perform extensive ablation analysis on ImageNet,  validating the effectiveness and superiority of our method.  The proposed classification head is flexible and fits  for a variety of vision  transformer architectures, significantly improving them on challenging image  classification tasks.  What's more,  the proposed  classification head generalizes to language transformer architecture, performing much better than the conventional classification head on general language understanding tasks. 

{\small
\bibliographystyle{ieee_fullname}
\bibliography{egbib}
}

\clearpage

\begin{appendices}
	\noindent{\Large\bf Supplemetary}
	\vspace{4mm}

\definecolor{mypink1}{rgb}{0.858, 0.188, 0.478}
\definecolor{mypink2}{RGB}{219, 48, 122}
\definecolor{mypink3}{cmyk}{0, 0.7808, 0.4429, 0.1412}
\definecolor{Gray}{gray}{0.95}
\definecolor{LightCyan}{rgb}{0.88,1,1}

\newcommand{\ra}{\rand0.\arabic{rand}}

\titlecontents{section}[18pt]{\bf\addvspace{10pt}}%
{\contentslabel{1.8em}}%
{}%
{\titlerule*[0.3pc]{$\cdot$}\contentspage}%

\titlecontents{subsection}[60pt]{}%
{\contentslabel{2.5em}}%
{}%
{\titlerule*[0.3pc]{$\cdot$}\contentspage}%

\titlecontents{subsubsection}[70pt]{}%
{\contentslabel{2.6em}}%
{}%
{\titlerule*[0.3pc]{$\cdot$}\contentspage}%

%\newtheorem{proof}{Proof}

%\numberwithin{equation}{section}
\renewcommand{\theequation}{S-\arabic{equation}}
\renewcommand\thetable{S-\arabic{table}}
\renewcommand\thefigure{S-\arabic{figure}}
%\renewcommand\thesection{\arabic{section}}

%%%%%%%%% TITLE
\appendix
\setcounter{section}{0}
\setcounter{table}{0}
\setcounter{figure}{0}
\setcounter{equation}{0}
\setcounter{proposition}{0}
\renewcommand{\appendixname}{}
%%%%%%%%% ABSTRACT
%\begin{abstract}
\tableofcontents

\vspace{12pt}
\section{Architectures of  Our Family of SoT}\label{suppsection:SoT network}

Following~[\hyperref[reference:ViT-supp]{\textcolor{green}{S-7}}], our transformer architecture consists of a token embedding module, a backbone and a classification head, as shown in Tab.~\ref{table:architecture}. For token embedding, the original ViT only uses a  single linear projection of fixed image patches, failing to model local image information. To address this limitation,  we design a small, hierarchical  module based on off-the-shelf convolutions. Note  that our simple embedding module has proven to be very competitive, compared to the other variants, as shown in the main paper. 

\paragraph{Token embedding}  Our embedding module consists of a stem and a stack of three convolution blocks, gradually decreasing image resolution. The stem is a $3\times 3$ convolution followed by a max pooling of stride 2 (S2). The design of blocks is flexible, and we choose    ResNet block~[\hyperref[reference:He_2016_CVPR-supp]{\textcolor{green}{S-9}}], Inception block~[\hyperref[reference:szegedy2016rethinking-supp]{\textcolor{green}{S-24}}], DenseNet block~[\hyperref[reference:huang2017densely-supp]{\textcolor{green}{S-11}}] and Non-local block~[\hyperref[reference:wang2018non-supp]{\textcolor{green}{S-29}}], whose configurations are shown in Fig.~\ref{fig:blocks}. We halve the spatial size of feature maps for each block, and after the last block, we use a  1$\times$1 convolution to change the dimension of features such that the dimension is consistent with that of the backbone. For  an input image of $224\times 224$, our token embedding outputs $14\times 14$ spatial features, reshaped to a sequence of 196 vectors as word tokens.

\paragraph{Backbone of transformer} We build the backbone by stacking standard transformer blocks as ~[\hyperref[reference:ViT-supp]{\textcolor{green}{S-7}}], where each transformer block contains a multi-head self-attention (MSA) and a multi-layer perception (MLP). Throughout the backbone, the dimension of tokens (token size) remain unchanged.  In MSA, we specify the number of heads; the dimension of queries, keys and values can be determined accordingly. Each MLP contains two  fully-connected (FC) layers, where the dimension of hidden layer is increased (called MLP size).  

\paragraph{Classification head} The proposed classification head combines the classification token  and word tokens, where word tokens are aggregated by multi-headed global cross-covariance pooling (MGCrP) with structured normalization svPN. We keep the number of heads unchanged and vary the dimensions ($m$ and $n$) of two linear projections for SoT of different depths.

We develop a family of SoT, namely, a light-weight 12-layer SoT-Tiny,  a middle-sized 14-layer SoT-Small and a heavyweight 24-layer SoT-Base, and their configurations are shown in Tab.~\ref{table:architecture}. In addition, to make computationally feasible the ablation analysis  where extensive evaluations on ImageNet are involved, we also design a 7-layer SoT. It  shares the same configuration with SoT-Tiny,  but only has 7 layers and moreover, the downsampling of the last block in the token embedding module is removed.

\begin{table}[t]
	\begin{minipage}{1.0\linewidth}
		\centering
		\footnotesize
		\setlength{\tabcolsep}{4pt}
		\renewcommand\arraystretch{1.3}
		\begin{tabular}{l|l|ccc}
			\hline			
			&  & SoT-Tiny & SoT-Small & SoT-Base \\
			\hline
			\parbox{0.5in}{\vspace{1mm}Token \\embedding}& Token size & 240 & 384 & 528 \\
			\hline
			\multirow{3}{*}{Backbone} & Layers & 12 & 14 & 24 \\
			& MSA heads & 4 & 6 & 8 \\
			& MLP size & 600 & 1344 & 1584 \\
			\hline
			Classification &MGCrP heads & 6 & 6 & 6 \\
			head& MGCrP $(m,n)$ &  (14,14) & (24,24) & (38,38) \\
			\hline
			\multicolumn{2}{c|}{Parameters (M)}  & 7.7 & 26.9 & 76.8 \\
			\multicolumn{2}{c|}{FLOPs (G)} & 2.5 & 5.8  &  14.5\\
			
			\hline
			
		\end{tabular}%
		\caption{Architectures   of the proposed SoT networks.}\label{table:architecture}%
	\end{minipage}\hfill
	
	\vspace{12pt}
	
	\begin{minipage}{1.0\linewidth}
		% \centering
		% \includegraphics[width=80mm]{images/blocks.pdf}
		% \captionof{figure}{Illustration of different type of convolution blocks in our module for token embedding. Left top: $\mathrm{DenseNet\ block}$; right top:  $\mathrm{Non}$-$\mathrm{local\ block}$; left bottom: $\mathrm{ResNet\ block}$; right bottom:$\mathrm{Inception\ block}$.}
		% \label{fig:blocks}
		\begin{subfigure}[b]{0.45\textwidth}
			\centering
			\includegraphics[height=1.75in]{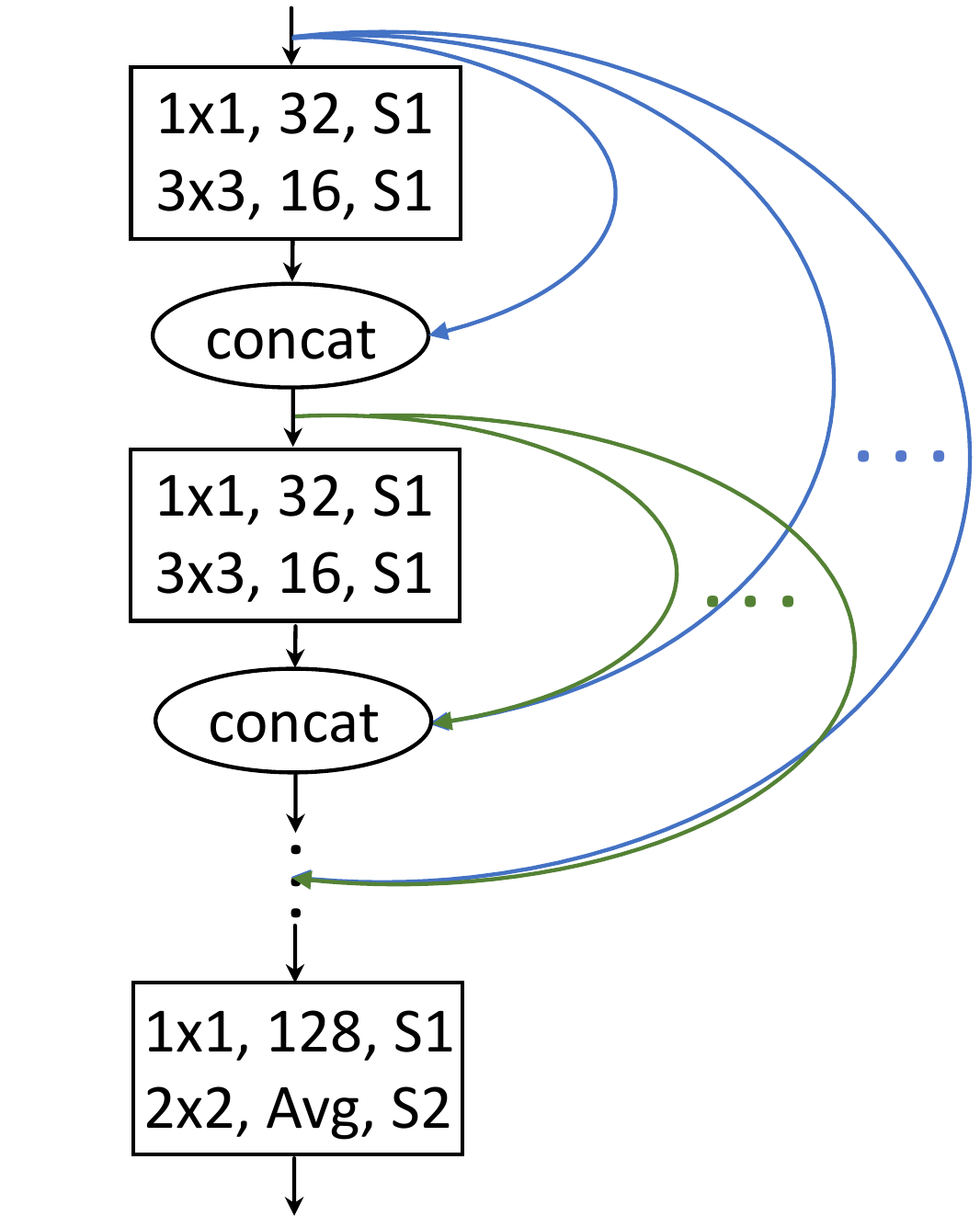}
			\caption{DenseNet block}
			\label{subfigure:DenseNet_block}
		\end{subfigure}%
		~\hspace{3pt}
		\begin{subfigure}[b]{0.4\textwidth}
			\centering
			\includegraphics[height=1.4in]{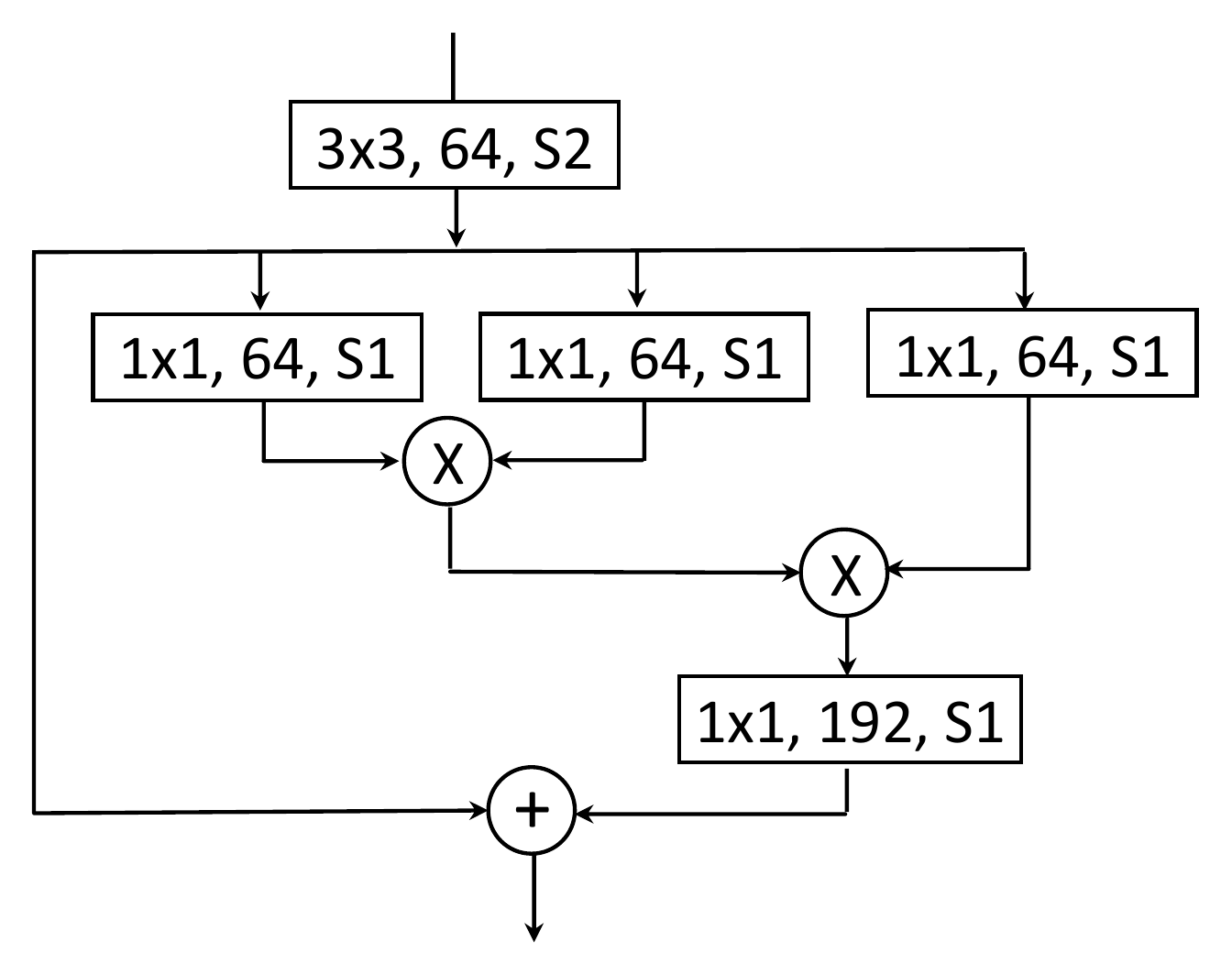}
			\caption{Non-local block}
			\label{subfigure:Non-local_block}
		\end{subfigure}
		
		\begin{subfigure}[b]{0.3\textwidth}
			\centering
			\includegraphics[height=1.35in]{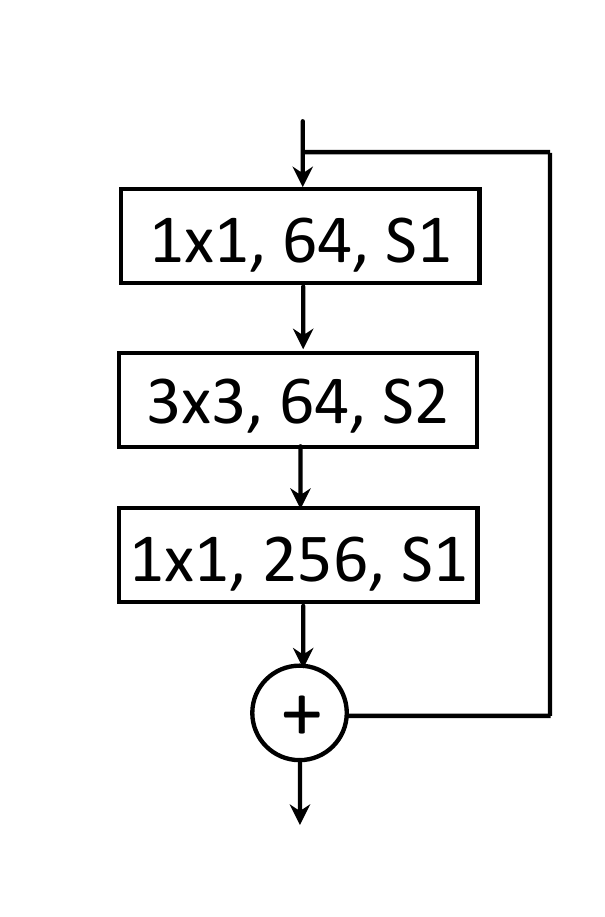}
			\caption{ResNet block}
			\label{subfigure:ResNet_block}
		\end{subfigure}%
		~\hspace{3pt}
		\begin{subfigure}[b]{0.60\textwidth}
			\centering
			\includegraphics[height=1.4in]{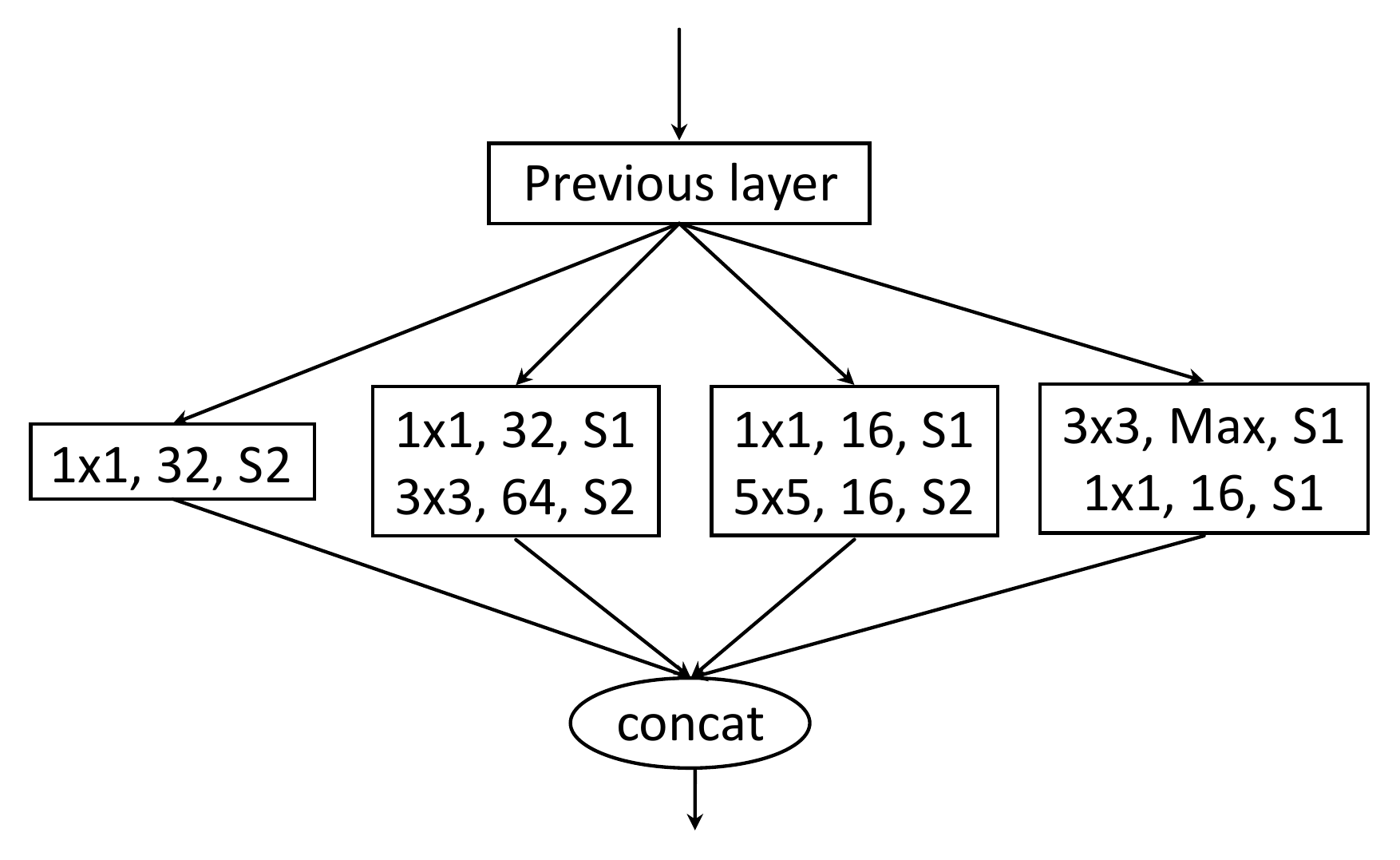}
			\caption{Inception block}
			\label{subfigure:Inception_block}
		\end{subfigure}	
		\captionof{figure}{Illustration of different type of convolution blocks in our  token embedding module. S1/2: Stride 1/2; Avg: average pooling; Max: max pooling.}
		\label{fig:blocks}
	\end{minipage}
\end{table}

\section{Proof of Proposition 1 }\label{suppsection:Proof of Proposition}

\begin{proposition}\label{proposition1}   
	\textit{
		Given $\mathbf{u}_{i}$ and $\mathbf{v}_{i}$ where $i=1,\ldots, k-1$,  let us consider  the objective 
		\begin{align}\label{equ:kth-cross-cov}
			\max_{\|\mathbf{u}\|=\|\mathbf{v}\|=1} \;&\mathbf{u}^{T}\mathbf{X}\mathbf{Y}^{T}\mathbf{v}\\
			\mathrm{s.t.}\;\;&\mathbf{u}_{i}^{T}\mathbf{u}=0, \;\mathbf{v}_{i}^{T}\mathbf{v}=0, \;i<k.\nonumber
		\end{align} 
		By inductively optimizing~(\ref{equ:kth-cross-cov}) for $k=1,\ldots, \min(m,n)$, we can obtain,  in decreasing order,  the $k$-th largest cross-covariance $\mathbf{u}_{k}^{T}\mathbf{X}\mathbf{Y}^{T}\mathbf{v}_{k}$, which  is equal to the $k$-th singular value  $\lambda_{k}$ of $\mathbf{Q}$ while $\mathbf{u}_{k}$ and $\mathbf{v}_{k}$ are the corresponding  left and right singular vectors, respectively.
	}
\end{proposition}

\begin{proof}
	We  prove this proposition using mathematical induction. Note that  $\mathbf{Q}=\mathbf{X}\mathbf{Y}^{T}$. For convenience, we let  $R(\mathbf{Q},\mathbf{u},\mathbf{v})=\mathbf{u}^{T}\mathbf{X}\mathbf{Y}^{T}\mathbf{v}$.

	\vspace{6pt}\noindent\textbf{Initial case }  First, let us consider the initial case of objective~(\ref{equ:kth-cross-cov}) for which  $k=1$,  i.e., $\max_{\|\mathbf{u}\|=\|\mathbf{v}\|=1}R(\mathbf{Q},\mathbf{u},\mathbf{v})$. Note  that $\|\mathbf{u}\|=1$ is equivalent to $\|\mathbf{u}\|^2=\mathbf{u}^{T}\mathbf{u}=1$. The Lagrange function associated with the objective~(\ref{equ:kth-cross-cov}) is 
	\begin{align}\label{equ:lagrange_1st}
		\mathcal{L}(\mathbf{u},\mathbf{v},\gamma,\beta)=R(\mathbf{Q}, \mathbf{u}, \mathbf{v})-\dfrac{\gamma}{2}(\mathbf{u}^{T}\mathbf{u}-1)-\dfrac{\beta}{2} (\mathbf{v}^{T}\mathbf{v}-1).
	\end{align}
	We calculate the gradient $\triangledown\mathcal{L}=\Big[\frac{\partial \mathcal{L}}{\partial \mathbf{u}},\frac{\partial \mathcal{L}}{\partial \mathbf{v}},\frac{\partial \mathcal{L}}{\partial \gamma},\frac{\partial \mathcal{L}}{\partial \beta}\Big]$,  where $\frac{\partial \mathcal{L}}{\partial \mathbf{u}}$ is the partial derivatives of  $\mathcal{L}$ with respect to $\mathbf{u}$, and set it to be zero. After some manipulations, we have
	\begin{align}\label{equ: a set of_first}
		\mathbf{Q}\mathbf{v}-\gamma \mathbf{u}=0,\\
		\mathbf{Q}^{T}\mathbf{u}-\beta \mathbf{v}=0,\nonumber\\
		\mathbf{u}^{T}\mathbf{u}-1=0,\nonumber\\
		\mathbf{v}^{T}\mathbf{v}-1=0.\nonumber
	\end{align}
	The last two equations simply produce the constraints, i.e., $\|\mathbf{u}\|=1$ and $\|\mathbf{v}\|=1$.  We left multiple  the first and second equation by $\mathbf{u}^{T}$ and $\mathbf{v}^{T}$, respectively. Then we can obtain 
	\begin{align*}
		\gamma=\beta=R(\mathbf{Q},\mathbf{u},\mathbf{v}), 
	\end{align*} 
	as  $\mathbf{u}^{T}\mathbf{Q}\mathbf{v}=\mathbf{v}^{T}\mathbf{Q}^{T}\mathbf{u}$. Therefore, we have  
	\begin{align}\label{equ: a pair of left and right SV}
		\mathbf{Q}\mathbf{v}=\gamma \mathbf{u},\\
		\mathbf{Q}^{T}\mathbf{u}=\gamma \mathbf{v}. \nonumber
	\end{align}
	According  to the property of SVD~[\hyperref[reference:Golub_4th]{\textcolor{green}{S-8}\textcolor{black}{, Chap.2.4}}], we know that $\mathbf{u}$ and $\mathbf{v}$ which satisfy~\ref{equ: a pair of left and right SV}  are respectively left and right singular vectors  of $\mathbf{Q}$ with $\gamma$ being the singular value. Obviously, $R(\mathbf{Q},\mathbf{u},\mathbf{v})$ achieves its maximum   when it is  equal to the largest singular value.  Therefore, by optimizing the objective~(\ref{equ:kth-cross-cov}) with $k=1$, we obtain  vectors $\mathbf{u}_{1}$ and $\mathbf{v}_{1}$ which are respectively the left and right singular vectors corresponding to the largest singular value $\lambda_{1}=R(\mathbf{Q},\mathbf{u}_{1},\mathbf{v}_{1})$.

	\vspace{6pt}\noindent\textbf{Inductive step} Next, we optimize the objective~(\ref{equ:kth-cross-cov}) for $k>1$. Suppose  the statement   holds for $i<k$. That is, for any $i$, $\mathbf{u}_{i}$ and $\mathbf{v}_{i}$, which maximize $R(\mathbf{Q},\mathbf{u},\mathbf{v})$ while satisfying the constraints $\|\mathbf{u}_{i}\|=\|\mathbf{v}_{i}\|=1$ and $\mathbf{u}_{i}^{T}{\mathbf{u}_{i'}}=0, \mathbf{v}_{i}^{T}{\mathbf{v}_{i'}}=0, i'<i$, are the singular vectors corresponding to the $i\mathrm{th}$ largest singular values $\lambda_{i}$. Obviously 
	\begin{align*}
		\underbrace{R(\mathbf{Q},\mathbf{u}_{1},\mathbf{v}_{1})}_{\lambda_{1}}\geq\ldots\geq \underbrace{R(\mathbf{Q},\mathbf{u}_{k-1},\mathbf{v}_{k-1})}_{\lambda_{k-1}}.
	\end{align*}
	
	Now, let us prove the statement holds for the case $k$. The   Lagrange function of  the objective (\ref{equ:kth-cross-cov}) is 
	\begin{align}\label{equ:lagrange_kth}
		\mathcal{L}(&\mathbf{u},\mathbf{v},\gamma,\beta,\tau_{i},\delta_{i})=R(\mathbf{Q},\mathbf{u},\mathbf{v})-\dfrac{\gamma}{2}(\mathbf{u}^{T}\mathbf{u}-1)\\
		&-\dfrac{\beta}{2} (\mathbf{v}^{T}\mathbf{v}-1)-\sum\nolimits_{i=1}^{k-1}\tau_{i}\mathbf{u}^{T}\mathbf{u}_{i}-\sum\nolimits_{i=1}^{k-1}\delta_{i}\mathbf{v}^{T}\mathbf{v}_{i}.\nonumber
	\end{align}
	We calculate the gradient of the Lagrange function $\triangledown\mathcal{L}=\Big[\frac{\partial \mathcal{L}}{\partial \mathbf{u}},\frac{\partial \mathcal{L}}{\partial \mathbf{v}},\frac{\partial \mathcal{L}}{\partial \gamma},\frac{\partial \mathcal{L}}{\partial \beta},\frac{\partial \mathcal{L}}{\partial \tau_{1}},\ldots, \frac{\partial \mathcal{L}}{\partial \tau_{k-1}},\frac{\partial \mathcal{L}}{\partial \delta_{1}},\ldots, \frac{\partial \mathcal{L}}{\partial \delta_{k-1}}\Big]$. By setting $\triangledown\mathcal{L}$ to be zero, we obtain a set of equations
	\begin{align}\label{equ: a set of_kth}
		\mathbf{Q}\mathbf{v}-\gamma \mathbf{u}-\sum\nolimits_{i=1}^{k-1}\tau_{i}\mathbf{u}_{i}=0,\\
		\mathbf{Q}^{T}\mathbf{u}-\beta \mathbf{v}-\sum\nolimits_{i=1}^{k-1}\delta_{i}\mathbf{v}_{i}=0,\nonumber\\
		\mathbf{u}^{T}\mathbf{u}-1=0,\nonumber\\
		\mathbf{v}^{T}\mathbf{v}-1=0,\nonumber\\
		\mathbf{u}^{T}\mathbf{u}_{i}=0,\; i=1,\ldots, k-1,\nonumber\\
		\mathbf{v}^{T}\mathbf{v}_{i}=0,\; i=1,\ldots, k-1.\nonumber
	\end{align}
	The third to  last equations are simply constraints of the maximization problem~(\ref{equ:kth-cross-cov}). We left multiply $\mathbf{u}^{T}$ (resp., $\mathbf{v}^{T}$) the first (resp., second) equation, and we can obtain  $\mathbf{u}^{T}\mathbf{Q}\mathbf{v}=\gamma$ (resp., $\mathbf{\mathbf{v}}^{T}\mathbf{Q}^{T}\mathbf{u}=\beta$), by noting that $\mathbf{u}^{T}\mathbf{u}_{i}=0$  (resp., $\mathbf{v}^{T}\mathbf{v}_{i}=0$) $\mathbf{u}$  for $i<k$.  Therefore, we have $\gamma=\beta=R(\mathbf{Q},\mathbf{u},\mathbf{v})$.  
	
	Subsequently, we will show that $\tau_{j}=0, j<k$.  We left multiply the first equation  by $\mathbf{u}_{j}^{T}, j<k$. We recall that $\mathbf{u}_{j}$ is orthogonal to $\mathbf{u}_{j'}$ for $j'\neq j$ and to $\mathbf{u}$, and then can derive  
	\begin{align*}
		\tau_{j}=\mathbf{u}_{j}^{T}\mathbf{Q}\mathbf{v}.
	\end{align*}
	As $\mathbf{u}_{j}$ is the left singular value of $\mathbf{Q}$, we know $\mathbf{Q}^{T}\mathbf{u}_{j}=\lambda_{j}\mathbf{v}_{j}$, i.e., $\mathbf{u}_{j}^{T}\mathbf{Q}=\lambda_{j}\mathbf{v}_{j}$. 
	Therefore we have  
	\begin{align*}
		\tau_{j}=\mathbf{u}_{j}^{T}\mathbf{Q}\mathbf{v}=\lambda_{j}\mathbf{v}_{j}^{T}\mathbf{v}=0.
	\end{align*}
	Here we make use of the fact that $\mathbf{v}_{j}$ is orthogonal to $\mathbf{v}$. In a similar manner, we left multiply the second equation by $\mathbf{v}_{j}^{T}, j<k$, and then we can derive $\delta_{j}=0$.  Up to this point, we   know $\mathbf{u}$ and $\mathbf{v}$  for which the objective~(\ref{equ:kth-cross-cov}) is maximized satisfy the following pair of equations 
	\begin{align}\label{equ: a pair of left and right SV--induction}
		\mathbf{Q}\mathbf{v}=\gamma \mathbf{u},\\
		\mathbf{Q}^{T}\mathbf{u}=\gamma \mathbf{v}. \nonumber
	\end{align}
	Again, according to the property of SVD, we know $\mathbf{u}$ and $\mathbf{v}$ are left and right singular values of $\mathbf{Q}$ and $\gamma$ is the corresponding singular value. Obviously,  $\gamma$ achieves the maximum when it equals the $k$-$\mathrm{th}$ largest singular value.  This concludes our proof. 
\end{proof}

As far as we know,  the statement as described in Proposition~\ref{proposition1} appeared early in~[\hyperref[reference:MCA_SVD_Climate-supp]{\textcolor{green}{S-2}}] and later in~[\hyperref[reference:book_statistical_1999-supp]{\textcolor{green}{S-27}\textcolor{black}{, Chap. 14.1.7}}], among others. However, we fail to find any formal proof of this statement; hence, we provide  a  proof here. It is worth mentioning that this statement is closely related to  but different from canonical correlation analysis (CCA). For detailed theory  on CCA, one may refer to~[\hyperref[reference:10.1145/3136624-supp]{\textcolor{green}{S-26}}].

\section{Backpropagation of svPN via SVD}\label{suppsection:Backpropagation}

% We give the backpropagation formulas for the exact singular value power normalization ( $\mathrm{svPN}$). %and for the approximate one (i.e., $\mathrm{\widehat{sv}PN}$). 

% \subsection{Backpropagation of svPN}\label{subsection:exact svPN layer}%$\mathrm{svPN}$}  

Let $\mathbf{Q}=	\mathbf{U}\mathrm{diag}(\lambda_{i})\mathbf{V}^{T}$ be the SVD  of $\mathbf{Q}$. The forward propagation of our normalization $\tilde{\mathbf{Q}}\stackrel{\vartriangle}{=}\mathrm{svPN}(\mathbf{Q})$ can be described in two consecutive steps as follows:
\begin{align}\label{equ:svPN}
\mathbf{Q}\stackrel{\;\mathrm{SVD}\;}{\longrightarrow}\mathbf{U}\mathrm{diag}(\lambda_{i})\mathbf{V}^{T}\stackrel{\mathrm{power}\;}{\longrightarrow}\mathbf{U}\mathrm{diag}(\lambda_{i}^{\alpha})\mathbf{V}^{T}
\end{align}
The associated backward propagations are  not that straightforward  as the structured,   nonlinear matrix operations are involved.  Suppose  $l$ is the network loss function. Let us denote $\mathbf{A}_{\mathrm{sym}}=\frac{1}{2}(\mathbf{A}+\mathbf{A}^{T})$, $\mathbf{\Lambda}=\mathrm{diag}(\lambda_{i})$,   and $\mathbf{A}_{\mathrm{diag}}$ being a matrix setting the off-diagonals of $\mathbf{A}$ to zero. Based on the theory of matrix backpropagation~[\hyperref[reference:Ionescu_arXiv15-supp]{\textcolor{green}{S-12}}], we can derive the gradients relative to $\mathrm{svPN}$ via SVD, which are given in the following corollary. 

\begin{corollary}\label{corollary:backpropagation of svPN}
Suppose we have  $\dfrac{\partial l}{\partial \tilde{\mathbf{Q}}}$ from the succeeding layer. The gradient involved in the first step of~(\ref{equ:svPN}) is  
\begin{align*}%\label{equ:partial derivative of SVD}
	&\dfrac{\partial l}{\partial \mathbf{Q}}=\dfrac{\partial l}{\partial \mathbf{U}}\mathbf{\Lambda}^{-1}\mathbf{V}^{T}+\mathbf{U}\Big(\dfrac{\partial l}{\partial \mathbf{\Lambda}}-\mathbf{U}^{T}\dfrac{\partial l}{\partial \mathbf{U}}\mathbf{\Lambda}^{-1}\Big)_{\mathrm{diag}}\mathbf{V}^{T}\\
	&+2\mathbf{U}\mathbf{\Lambda}\Big(\mathbf{K}^{T}\circ \Big( \mathbf{V}^{T}\Big( \dfrac{\partial l}{\partial \mathbf{V}}-\mathbf{V}\mathbf{\Lambda}^{-1} \Big(\dfrac{\partial l}{\partial \mathbf{U}}\Big)^{T}\mathbf{U}\mathbf{\Lambda} \Big)\Big) \Big)_{\mathrm{sym}}\mathbf{V}^{T}
\end{align*}
where  $K_{ij}=(\lambda_{i}^2-\lambda_{j}^2)^{-1}$  if $\lambda_{i}\neq \lambda_{j} $ and $K_{ij}=0$ otherwise,  and  $\circ$ denotes Hadamard  product. The partial derivatives with respect to the second step of~(\ref{equ:svPN}) are
\begin{align*}
	\dfrac{\partial l}{\partial \mathbf{U}}&=\dfrac{\partial l}{\partial \tilde{\mathbf{Q}}}\mathbf{V}\mathbf{\Lambda}^{\alpha},\\
	\dfrac{\partial l}{\partial \mathbf{V}}&=\Big(\dfrac{\partial l}{\partial \tilde{\mathbf{Q}}}\Big)^{T}\mathbf{U}\mathbf{\Lambda}^{\alpha},\\
	\dfrac{\partial l}{\partial \mathbf{\Lambda}}&=\alpha\mathbf{\Lambda}^{\alpha-1}\mathbf{U}^{T}\dfrac{\partial l}{\partial \tilde{\mathbf{Q}}}\mathbf{V}.
\end{align*}
\end{corollary}

%%%%%%%%% BODY TEXT
\section{Detailed Experimental Settings for CV and NLP Tasks}\label{suppsection:detailed}

\subsection{Benchmark Description}

\subsubsection{Benchmarks Used in CV}

\vspace{6pt}\noindent\textbf{ImageNet}  Our experiments are mainly conducted  on ILSVRC  ImageNet 2012 image classification  benchmark~[\hyperref[reference:ILSVRC15-supp]{\textcolor{green}{S-21}},\hyperref[reference:imagenet_cvpr09-supp]{\textcolor{green}{S-5}}], which contains  1K classes with 1.28M images for training and  50K images for validation. Note that as test images are not publicly available, the common practice is to adopt the validation images for testing~[\hyperref[reference:T2T_ICCV21-supp]{\textcolor{green}{S-34}},\hyperref[reference:DeiT_ICML-supp]{\textcolor{green}{S-25}},\hyperref[reference:PSViT_ICCV21-supp]{\textcolor{green}{S-35}}]. We train the transformer models from scratch on the training set and  report  top-1 accuracy on the validation set.

\begin{figure}[t]
	\begin{center}
		\includegraphics[width=1.0\linewidth]{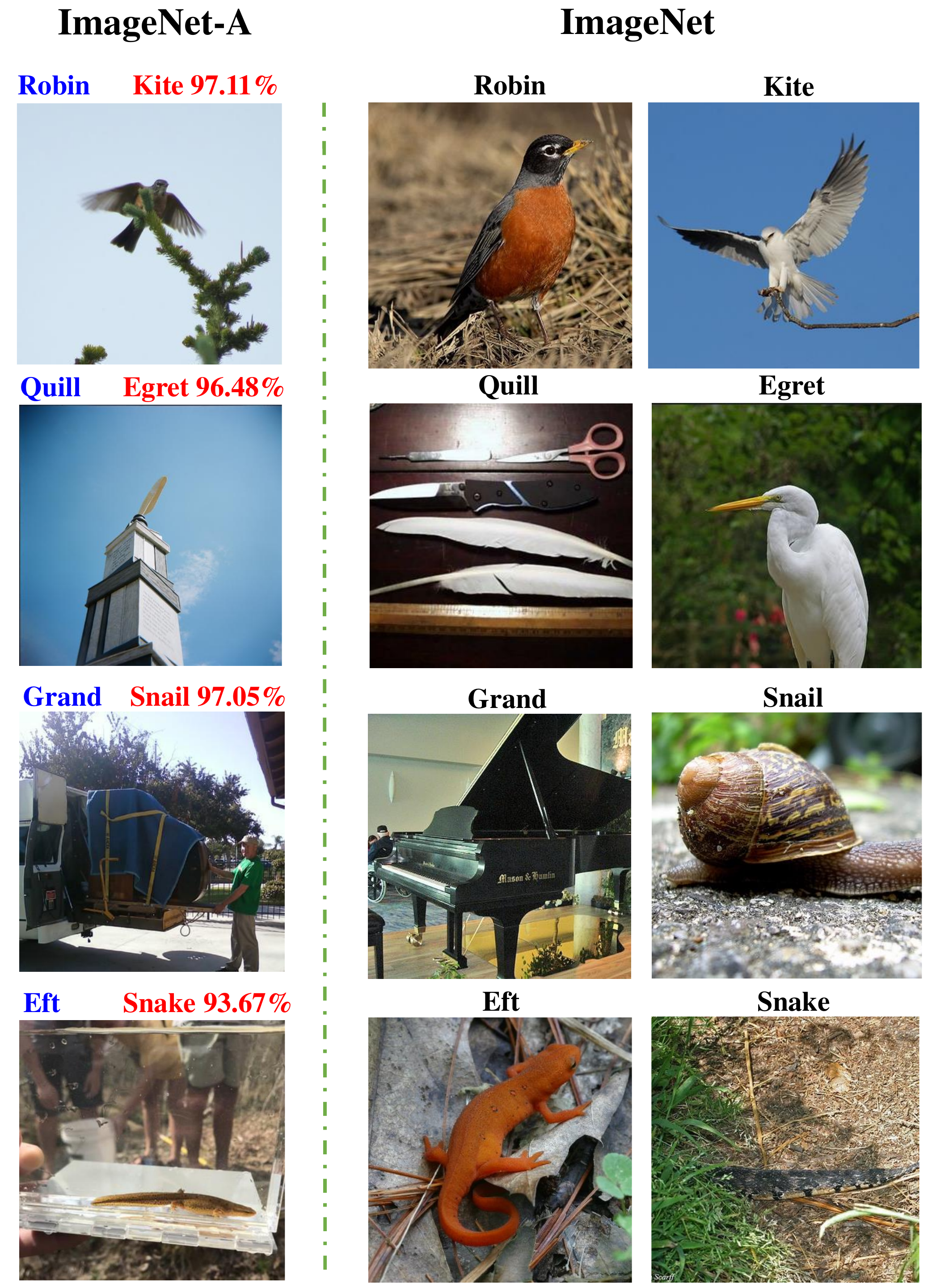}
	\end{center}
	\caption{Examples of adversarial  images from  ImageNet-A (1st column). The {\color{blue}{blue}} text  is the test class, and the {\color{red}{red}} text is the false  prediction and the score produced by a DeiT-Small model. The 2nd and 3rd columns show the images of the corresponding classes from ImageNet. It can be seen that the images from ImageNet-A are highly adversarial, whose distribution  deviates much from that of ImageNet. On the challenging ImageNet-A,  the proposed method can substantially improve state-of-the-art vision transformers (see Sec.~\hyperref[subsection:experiment-comparion with state-of-the-art]{\textcolor{red}{4.2.3}} in the main paper).   }
	\label{fig:ImageNet-A}
	%\vspace{-6pt}
\end{figure}

\vspace{6pt}\noindent\textbf{ImageNet-A} The ImageNet-A~[\hyperref[reference:ImageNet-A-supp]{\textcolor{green}{S-10}}] is a hard ImageNet test set of  real-world adversarial  images with adversarial filtration.  It   contains 7,500 natural images from 200 classes which cover most broad categories spanned by ImageNet. This dataset has heterogeneous and varied distribution shift from ImageNet. ImageNet-A is far more challenging than the original ImageNet validation set (and test set). For example,  DeiT-Small model achieves only a top-1 accuracy of 18.9\%  on ImageNet-A against 79.8\% on ImageNet validation set.  Fig.~\ref{fig:ImageNet-A} (1st column) shows four examples from ImageNet-A, in which an object ({\color{blue}{blue}}  text) of some class is mistakenly identified as that of another class with high confidence score  (in {\color{
red}{red}}); for  contrast, the 2nd and 3rd columns show the images of the corresponding classes in ImageNet. It can be seen that images of ImageNet-A is highly adversarial and its distribution deviates much from that of ImageNet. Notably, on the challenging ImageNet-A,  the proposed method can substantially improve state-of-the-art vision transformers (see Sec.~\hyperref[subsection:experiment-comparion with state-of-the-art]{\textcolor{red}{4.2.3}} in the main paper).

\subsubsection{Benchmarks Used in NLP}

\vspace{6pt}\noindent\textbf{CoLA} \quad The goal of the Corpus of Linguistic Acceptability task~[\hyperref[reference:warstadt2018neural-supp]{\textcolor{green}{S-30}}]  is to judge whether a sentence is grammatical or not. It can be formulated as a binary single-sentence classification problem.  The dataset contains  10,657 English sentences which are labeled as grammatical or ungrammatical, and these sentences are split into training (8,551)/development (1,043)/test (1,063) sets.

\vspace{6pt}\noindent\textbf{RTE} \quad Given a pair of text fragments, denoted by (``Text'', ``Hypothesis''), Recognizing Textual Entailment~[\hyperref[reference:bentivogli2009fifth-supp]{\textcolor{green}{S-1}}] aims to determine whether the ``Text''  entails ``Hypothesis''.  This task can be converted into a binary entailment classification task. The dataset of  RTE consists of  5,767 examples,  among which the training set contains 2,490 examples, while the development set  and test set contain  3,000 and 277 examples, respectively. 

\begin{figure*}[t]
	\centering
	\begin{subfigure}[b]{0.5\textwidth}
		\centering
		\includegraphics[height=3.0in]{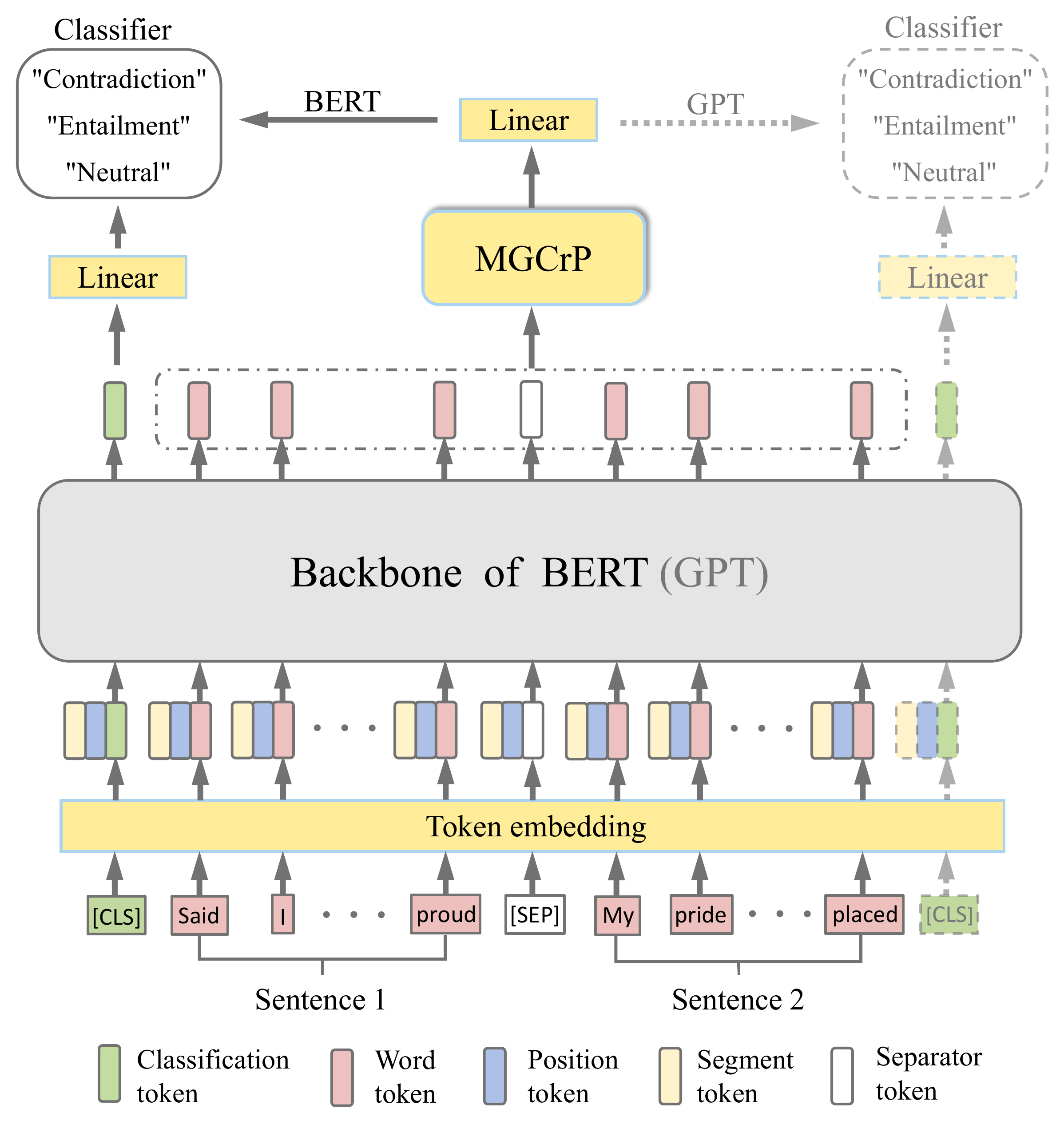}
		\caption{Sentence-pair classification task (RTE, MNLI and QNLI).}
		\label{subfigure:nlp_a}
	\end{subfigure}%
	\begin{subfigure}[b]{0.5\textwidth}
		\centering
		\includegraphics[height=3.0in]{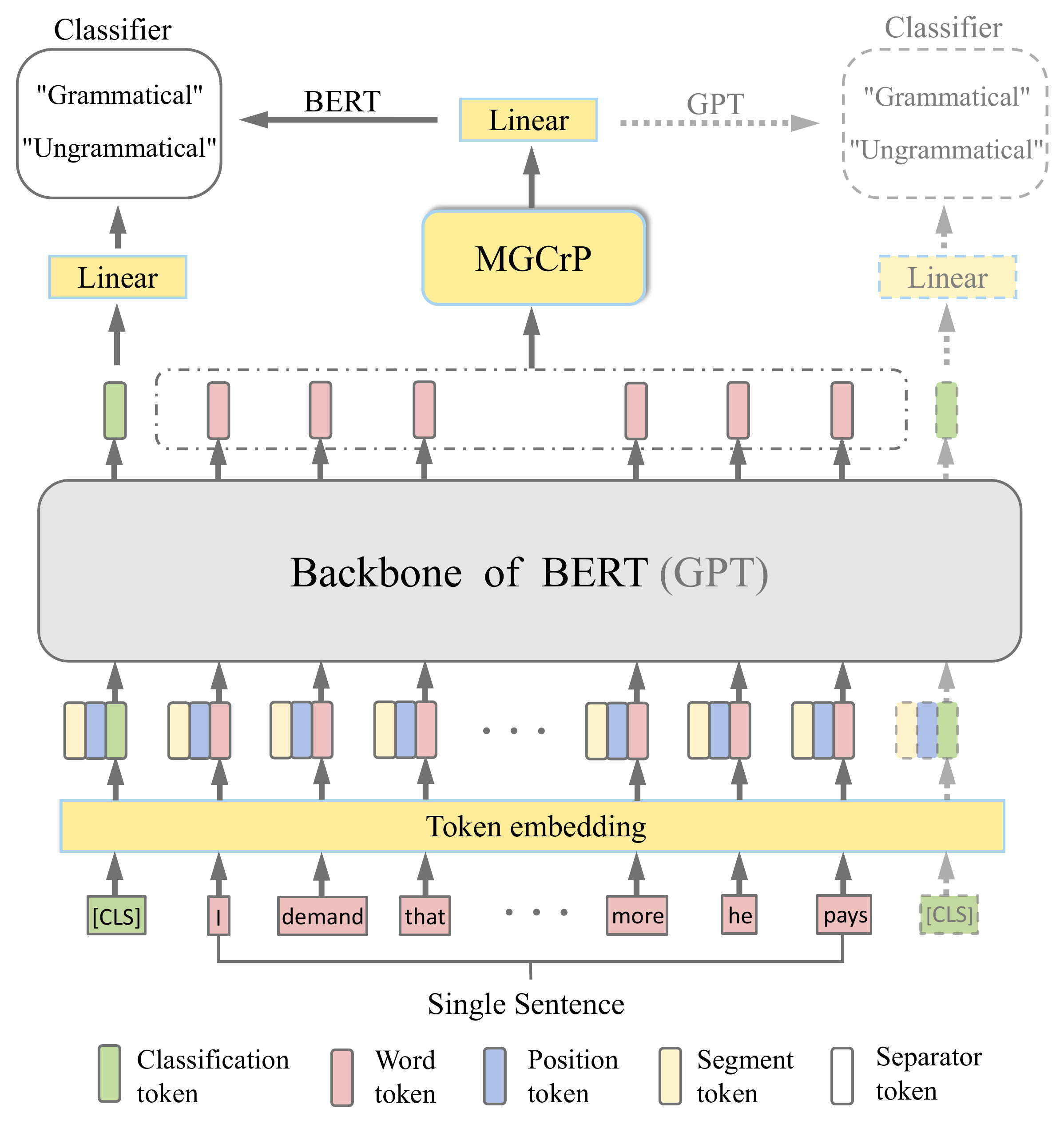}
		\caption{Single-sentence classification task (CoLA)}
		\label{subfigure:nlp_b}
	\end{subfigure}
	\caption{Diagrams of fine-tuning BERT and GPT on  downstream NLP tasks, formulated as either (a) sentence-pair classification (RTE, MNLI and QNLI), or (b) single-sentence classification (CoLA). Note that for BERT \& its variants, the classification token [CLS] is always in the first place of the sequence, while it is at the end for GPT whose illustration fades;  besides, GPT does not use segment embedding. }
	\label{fig:fine-tuning NLP}
\end{figure*}

\begin{table}[t]
	\centering
	\footnotesize

	\begin{minipage}{1\linewidth}
		\begin{subtable}{1\linewidth}
			\centering
			\setlength{\tabcolsep}{6pt}
			\renewcommand\arraystretch{1.2}
			\begin{tabular}{c|c|ccc}
				\hline
				\multicolumn{2}{c|}{Models} & SoT-Tiny&SoT-Small& SoT-Base \\
				\hline
				
				\multicolumn{2}{c|}{Batch size} & 1024 & 1024 & 512  \\
				\hline
				\multicolumn{2}{c|}{Optimizer} & AdamW &AdamW& AdamW\\
				% AdamW & \multirow{2}{*}{momentum} & 0.9 & 0.9 & 0.9 \\
				% Optimizer & & & &  \\
				
				\hline
				% \multirow{3}{*}{Warmup} & Epochs  &5 &  5&  5\\
				% & LR & 1e-6& 1e-6 & 1e-6 \\
				% & Scheduler &Linear & Linear& Linear\\
				% \hline
				
				\multicolumn{2}{c|}{Epochs}& 310 & 310 & 310 \\
				\multicolumn{2}{c|}{Base LR} & 1e-3& 1e-3 & 5e-4\\
				\multicolumn{2}{c|}{Final LR} & 1e-5 & 1e-5& 1e-5\\
				\multicolumn{2}{c|}{Scheduler} & cosine & cosine & cosine\\
				\multicolumn{2}{c|}{Weight decay} & 0.03 & 0.03  & 0.065 \\
				\hline
				\multicolumn{2}{c|}{Label smoothing }& 0.1 & 0.1 & 0.1\\
				\multicolumn{2}{c|}{Mixup prob.} & 0.8& 0.8 & 0.8\\
				\multicolumn{2}{c|}{Cutmix prob.} & 1.0 & 1.0& 1.0\\
				\multicolumn{2}{c|}{Erasing prob.} &0.25&0.25 & 0.25 \\
				\multicolumn{2}{c|}{RandAugment} & 9/0.5& 9/0.5 & 9/0.5\\
				\hline
				% \multicolumn{2}{c|}{MGCrP repr. size }&1k&3k&8k\\
				\multicolumn{2}{c|}{MGCrP dropout }&0.0&0.5&0.7 \\
				\hline

			\end{tabular}%
			%\caption{Training from scratch for image classification on ImageNet.}\label{table:cv_hyper}%
			
		\end{subtable}
	\end{minipage}
	
	\caption{Hyper-parameters for image classification.}
	\label{table:hyper-parameters-image}	
\end{table}

\vspace{6pt}\noindent\textbf{MNLI} \quad Similar to RTE, Multi-Genre Natural Language Inference~[\hyperref[reference:williams2018broad-supp]{\textcolor{green}{S-31}}]  is also concerned with judgement of entailment.  Given a pair of sentences, the task is to predict whether the ``Text''  entails the ``Hypothesis''  (entailment), contradicts the ``Hypothesis''  (contradiction), or neither (neutral). As such, this task can be formulated as a three-way classification problem.  MNLI is a large-scale dataset, consisting  of 432,702 examples, in which  392,702 examples belong to the training set,  20,000 examples belong to the  development set and the remaining 20,000 examples are in the test set.

\vspace{6pt}\noindent\textbf{QNLI} \quad The Stanford Question Answering task~[\hyperref[reference:rajpurkar2016squad-supp]{\textcolor{green}{S-20}}] is  converted to a sentence-pair classification problem~[\hyperref[reference:wang2018glue-supp]{\textcolor{green}{S-28}}]. Given a pair of  sentences, the model needs to determine whether the sentence contains the answer to the question.  QNLI dataset contains 105K training examples, 5.4K development examples and 5.4K test examples.

\subsection{Training Strategy}

\subsubsection{Training   from Scratch  on CV Tasks}  As suggested in~[\hyperref[reference:ViT-supp]{\textcolor{green}{S-7}}], training of high-performance transformer models requires ultra large-scale datasets. Hence,  for training from scratch on ImageNet which is not that large, one often depends on extensive data augmentation and regularization methods~[\hyperref[reference:DeiT_ICML-supp]{\textcolor{green}{S-25}}], for which we mainly follow ~[\hyperref[reference:DeiT_ICML-supp]{\textcolor{green}{S-25}},\hyperref[reference:T2T_ICCV21-supp]{\textcolor{green}{S-34}},\hyperref[reference:PSViT_ICCV21-supp]{\textcolor{green}{S-35}}]. For data augmentation, besides  standard scale, color and flip jittering~[\hyperref[reference:Simonyan15-supp]{\textcolor{green}{S-14}},\hyperref[reference:He_2016_CVPR-supp]{\textcolor{green}{S-9}}] with default settings in PyTorch, we adopt randAugment~[\hyperref[reference:cubuk2020randaugment-supp]{\textcolor{green}{S-4}}] and random erasing~[\hyperref[reference:zhong2020random-supp]{\textcolor{green}{S-39}}]. For model regularization, we employ label smoothing~[\hyperref[reference:Szegedy_2015_CVPR-supp]{\textcolor{green}{S-23}}],   mixup~[\hyperref[reference:zhang2018mixup-supp]{\textcolor{green}{S-38}}] and  cutmix~[\hyperref[reference:yun2019cutmix-supp]{\textcolor{green}{S-36}}]. We use AdamW~[\hyperref[reference:loshchilov2018decoupled-supp]{\textcolor{green}{S-17}}] optimizer  with a learning rate warm up (5 epochs)  and cosine annealing scheduler. We adopt dropout for our MGCrP module.  The hyper-parameters involved in augmentation, regularization, optimization, etc., are summarized in Tab.~\ref{table:hyper-parameters-image}. Note that the 7-layer SoT used in the ablation study shares  the same hyper-parameters with SoT-Tiny.

\begin{table}[t]
	\centering
	\footnotesize
	\begin{minipage}{1.0\linewidth}
		\begin{subtable}{1.0\linewidth}
			\centering
			\setlength{\tabcolsep}{1.5pt}
			\renewcommand\arraystretch{1.5}
			\begin{tabular}{c|c|c|c|c|c}
				\hline
				\multicolumn{2}{c|}{Models}&GPT& BERT &SpanBERT&RoBERTa\\
				\hline
				\multicolumn{2}{c|}{Batch size}& \{32,64\}& \{24,64,96\} &32&\{16,32\} \\
				\hline
				\multirow{2}{*}{Adam}& $\beta_{1}$ & 0.9&0.9& 0.9 & 0.9\\
				\multirow{2}{*}{Optimizer}&$\beta_{2}$ & 0.999&0.999& 0.999 & 0.98\\
				&$\epsilon$ & 1e-8 & 1e-8 & 1e-6 & 1e-6 \\
				\hline
				\multicolumn{2}{c|}{Epochs}& \{10,15\}&10&\{10,30\} &\{10,20\}\\
				\multicolumn{2}{c|}{Base LR}& 6.25e-5&\{2e-5,3e-5\}&2e-5&\{1e-5,2e-5\}\\
				\multicolumn{2}{c|}{Final LR} & 0& 0&0& 0\\
				\multicolumn{2}{c|}{Scheduler} & linear & linear & linear & linear \\
				\multicolumn{2}{c|}{Weight decay} & 0.01 & 0 & 0.01 & 0.1 \\
				% \hline
				% Drop prob.&\{0.5,0.7\}&\{0.5,0.8\}&\{0.5,0.7\}&\{0.5,0.7\}  \\
				\hline
				% Epochs& \{10,15\}&10&\{10,30\} &\{10,20\}\\
				% \hline
				\multicolumn{2}{c|}{MGCrP repr. size}& \{4K,6K\}&\{1K,4K,5K\} &\{1K,4K,5K\}& \{1K,4K,5K\} 
				\\
				\multicolumn{2}{c|}{MGCrP dropout} &\{0.5,0.7\}&\{0.5,0.8\}&\{0.5,0.7\}&\{0.5,0.7\} \\
				\hline
			\end{tabular}%
		\end{subtable}
	\end{minipage}
	\caption{Hyper-parameters for  text classification.}
	\label{table:hyper-parameter text}	
\end{table}

\subsubsection{Fine-tuning on Downstream NLP Tasks} 

The illustration of fine-tuning BERT and GPT with our classification head can be seen in Fig~\ref{fig:fine-tuning NLP}.
The four NLP tasks are formulated as either sentence-pair classification task (RTE, MNLI and QNLI) or single-sentence classification task (CoLA).      For each task, we plug in the task-specific input and outputs into the transformer models and fine-tune  the whole networks in an end-to-end fashion. Following previous works~[\hyperref[reference:roberta-supp]{\textcolor{green}{S-16}},\hyperref[reference:GPT-1-supp]{\textcolor{green}{S-19}},\hyperref[reference:DBLP:conf/naacl/DevlinCLT19-supp]{\textcolor{green}{S-6}},\hyperref[reference:joshi2020spanbert-supp]{\textcolor{green}{S-13}}],  for each task we  fine-tune the  model on the training set while evaluating on the development set.

In the following, we  introduce the fine-tuning pipeline by  taking the sentence-pair classification with BERT model as an example.  As shown in Fig.~\ref{subfigure:nlp_a}, for a sentence-pair classification, a pair of sentences are concatenated into a single sequence with a special token ([$\mathrm{SEP}$]) separating them, and is then prepended by a classification token ([$\mathrm{CLS}$]). The input representation of every token is built by summing the word embedding (by e.g., WordPiece~[\hyperref[reference:WordPiece-supp]{\textcolor{green}{S-33}}]), segment embedding and position embedding. At the output, the token representations are fed into our proposed classification head,  in which we combine  classification token and word tokens.  Fig.~\ref{subfigure:nlp_b} shows the single-sentence classification task, which is similar to sentence-pair task except the input only involves one sentence. We mention that the [$\mathrm{CLS}$] is always in the first place of the token sequence for BERT, while in the last for GPT.

For GPT and BERT \& its variants,  most training strategies and  hyper-parameters in fine-tuning are the same as those in pre-training. We use Adam~[\hyperref[reference:Adam-supp]{\textcolor{green}{S-15}}] algorithm for model optimization. The learning rate is linearly warmed up over a number of steps to a peak value, and then linearly decayed to zero.   We mainly tune the batch size, learning rate and number of training epochs.   The optimal hyper-parameters are task-specific. Following~[\hyperref[reference:GPT-1-supp]{\textcolor{green}{S-19}},\hyperref[reference:DBLP:conf/naacl/DevlinCLT19-supp]{\textcolor{green}{S-6}}], we choose them from a small set of options; for example, for GPT, we select batch size from \{32,64\}. 
For our classification head, we use dropout for  MGCrP.  For simplicity,  we adopt single head for our MGCrP and select the representation (repr.) size from a set of values, e.g., \{4K,6K\} for GPT.
The detailed settings of the hyper-parameters  are summarized in Tab.~\ref{table:hyper-parameter text}.

Fine-tuning of BERT and GPT are implemented based on HuggingFace’s codebase~[\hyperref[reference:DBLP:journals/corr/abs-1910-03771-supp]{\textcolor{green}{S-32}}],  while that of RoBERTa is based on fairseq~[\hyperref[reference:ott2019fairseq-supp]{\textcolor{green}{S-18}}], an  open-source sequence modeling toolkit. We implement fine-tuning of SpanBERT using the code available at 
 \href{https://github.com/facebookresearch/SpanBERT}{official website of facebook}.  
The pretrained BERT model is downloaded from  \href{https://huggingface.co/models}{HuggingFace's website}, and  pretrained RoBERTa and SpanBERT models are both from  \href{http://dl.fbaipublicfiles.com/fairseq/models}{fairseq website}. The  pretrained GPT model is  available at \href{https://github.com/openai/finetune-transformer-lm/tree/master/model}{the official website of OpenAI }.

\begin{figure*}[t!]
	\centering
	\includegraphics[width=6.9in]{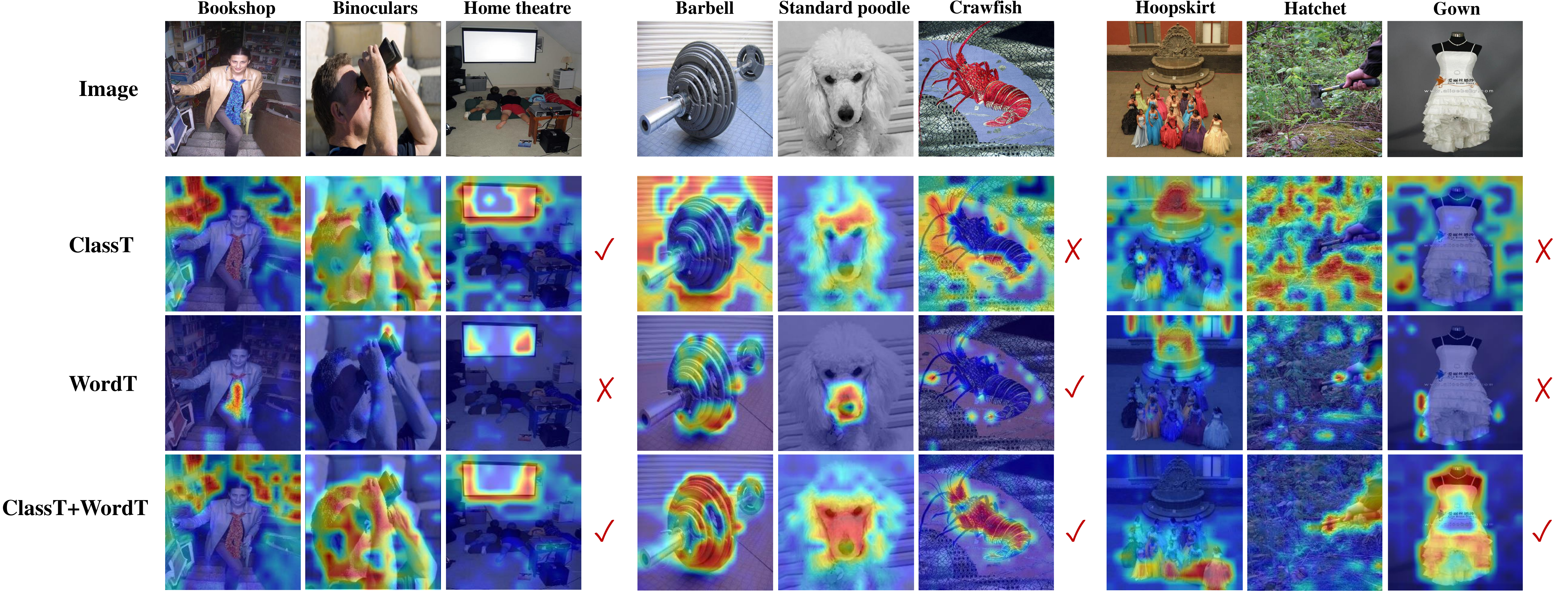}
	\caption{Visualizations of images on ImageNet validation set based on SoT-Tiny by using the Grad-CAM~[\hyperref[reference:selvaraju2017grad-supp]{\textcolor{green}{S-22}}]. We show the examples for which \textbf{ClassT}+\textbf{WordT}  predicts correctly, but   \textbf{ClassT}  or \textbf{WordT} fails. {\color{red}{\checkmark}}: correct prediction; {\color{red}{\xmark}}: incorrect prediction.}
	\label{figure:image_vis}
\end{figure*}

\vspace{12pt}
\begin{figure*}[t!]
	\centering
	\includegraphics[width=6.0in]{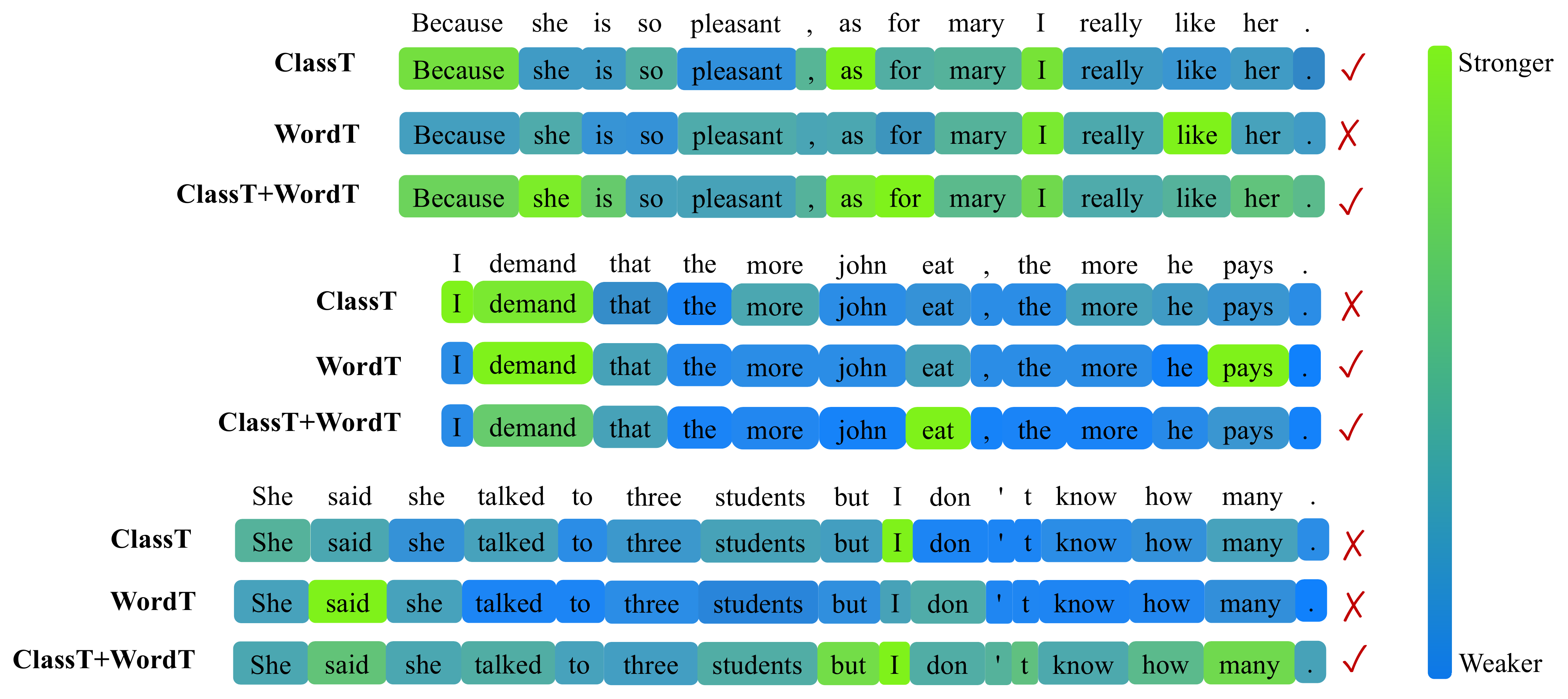}
	\caption{Visualization of the influence of each word for linguistic acceptability on the given English sentence. We adopt the BERT-base as the backbone and refer~[\hyperref[reference:Yun2021TransformerVis-supp]{\textcolor{green}{S-37}},\hyperref[reference:chefer2021transformer-supp]{\textcolor{green}{S-3}}] to obtain the results.  {\color{red}{\checkmark}}: correct prediction; {\color{red}{\xmark}}: incorrect prediction.}
	\label{figure:nlp_vis}
\end{figure*}

\section{Visualization for CV and NLP Tasks}

To further analyze the effectiveness of our proposed classification head, we make qualitative comparisons by visualizing the models for CV and NLP tasks. Specifically, SoT-Tiny and the BERT-base are used as the backbone models for CV and NLP tasks, respectively. For each model, we compare three variants as follows:
\begin{itemize}
	\item  \textbf{ClassT}: only classification token is used for classifier;
	\item \textbf{WordT}: only word tokens are used for classifier;
	\item \textbf{ClassT}+\textbf{WordT}: Both classification token and word tokens are used for classifier based on the  sum  scheme.
\end{itemize}

\subsection{Visualization for CV Model }  To visualize the models in CV task, we first train our  SoT-Tiny variants on ImageNet, and then adopt the Grad-CAM~[\hyperref[reference:selvaraju2017grad-supp]{\textcolor{green}{S-22}}] to obtain class activation map of each input image. As such, we can visualize the most important regions (i.e., regions of interest) for the final classification according to the gradient information. As illustrated in Fig.~\ref{figure:image_vis},  we show three kinds of scenarios, in which \textbf{ClassT}+\textbf{WordT} all makes correct predictions, i.e.,  (left panel) \textbf{ClassT} makes  correct predictions but \textbf{WordT} fails,  (middle panel) \textbf{WordT} predicts correctly  but \textbf{ClassT} does not; (right panel)  neither \textbf{ClassT} nor \textbf{WordT} predicts correctly.

From Fig.~\ref{figure:image_vis}, we have the following observations: 
(1) As classification token interacts with all word tokens across the network, it tends to  focus on the \textit{global} context of images, especially some messy backgrounds. Therefore, \textbf{ClassT} is more suitable for classifying the categories associated with the backgrounds and the whole context, e.g.,  “Bookshop”. (2) The word tokens  mainly correspond to local patches, so  \textbf{WordT} performs classification primarily based on some \textit{local} discriminative regions. As such, \textbf{WordT} has  better ability to classify the categories associated with local parts and subtle variations, e.g.,  “Standard poodle”. (3) Our  \textbf{ClassT}+\textbf{WordT} can make fully use of merits of both word tokens and classification token, which can focus on the most important regions for better classification by exploiting both local discriminative parts and global context information.

\subsection{Visualization for NLP Model } Similarly, we compare the visualization results of the BERT-base  under  three scenarios on the examples of CoLA.  The task is to judge whether an  English sentence is grammatical or not. We use  visualization methods proposed in~[\hyperref[reference:Yun2021TransformerVis-supp]{\textcolor{green}{S-37}},\hyperref[reference:chefer2021transformer-supp]{\textcolor{green}{S-3}}] to show the influence of each word in the final prediction. As shown in Fig.~\ref{figure:nlp_vis}, the \colorbox[RGB]{124,242,27}{green} denotes  stronger impact while the \colorbox[RGB]{13,119,234}{blue} implies  weaker one. 

All examples in Fig.~\ref{figure:nlp_vis} are ungrammatical. Overall, we can see  \textbf{ClassT}  inclines  to make predictions  from the whole sentence, such as the conjunction of two sub-sentences (e.g., ``Because.., as'') or the key global semantic word;  \textbf{WordT} tends to focus on local correctness of each sentence, ignoring  the global context. This observation is similar with the visualization results of CV model, demonstrating that the classification token and word tokens are highly complementary for both  CV and NLP tasks. Finally, the proposed \textbf{ClassT}+\textbf{WordT} can highlight all important words in sentence, including the subordinate clause, conjunction, etc., which can help to boost the performance of classification.

\makeatletter
\def\@bibitem#1{\item\if@filesw \immediate\write\@auxout
	{\string\bibcite{#1}{S-\the\value{\@listctr}}}\fi\ignorespaces}
\def\@biblabel#1{[S-{#1}]}
\makeatother

\phantomsection
\small{
\bibliographystylesupp{ieee_fullname}
%\bibliographysupp{egbib_supp}

}

\end{appendices}

\end{document}